\documentclass[10pt,twocolumn,letterpaper]{article}

\usepackage{iccv}
\usepackage[accsupp]{axessibility}
\usepackage{times}

\usepackage{epsfig}
\usepackage{graphicx}
\usepackage{amsmath}
\usepackage{amssymb}
\usepackage{arydshln}
\usepackage{enumitem}
\usepackage{subcaption}

\usepackage{microtype}
\usepackage{booktabs} 
\usepackage{multirow}
\usepackage{tikz}
\usepackage{dblfloatfix}
\usepackage{pgfplots}
\pgfplotsset{compat=1.16}
\usepgfplotslibrary{groupplots}
\usetikzlibrary{patterns}

\usetikzlibrary{arrows}
\usepackage{tkz-kiviat}
\usepackage{xfp}

\usepackage{graphbox}
\usepackage{booktabs}       
\usepackage{amsfonts}       
\usepackage{nicefrac}       
\usepackage{microtype}      
\usepackage{comment}
\usepackage{amsmath,amssymb} 
\usepackage{caption}
\usepackage{float}
\usepackage{algpseudocode}
\usepackage{algorithm}

\usepackage[pagebackref=true,breaklinks=true,letterpaper=true,colorlinks,bookmarks=false]{hyperref}
\usepackage{cleveref}

\usepackage{xcolor}
\hypersetup{
    colorlinks,
    linkcolor={red},
    citecolor={green}
}
\definecolor{citecolor}{HTML}{0071bc}
\definecolor{color_ao}{gray}{0.5}
\definecolor{color_our}{HTML}{78debc}
\definecolor{color_pre}{rgb}{0.52,0.59,0.69}
\definecolor{Gray}{gray}{0.9}
\definecolor{LighterGray}{gray}{0.93}
\definecolor{LightGrayForTableRule}{gray}{0.92}
\definecolor{DarkGray}{gray}{0.5}
\definecolor{Black}{rgb}{0.0, 0.0, 0.0}
\definecolor{NiceBlue}{rgb}{0.11764705882352941, 0.5647058823529412, 1.0}
\definecolor{NiceGreen}{HTML}{008080}
\usepackage{diagbox}
\definecolor{gray}{HTML}{b0aeae}
\definecolor{diffgray}{HTML}{878787}
\definecolor{NiceGray}{HTML}{696969}
\definecolor{applegreen}{rgb}{0.55, 0.71, 0.0}

\definecolor{demphcolor}{RGB}{144,144,144}
\newcommand{\demph}[1]{\textcolor{demphcolor}{#1}}

\newcommand\blfootnote[1]{%
  \begingroup
  \renewcommand\thefootnote{}\footnote{#1}%
  \addtocounter{footnote}{-1}%
  \endgroup
}

\usepackage{mathtools}
\usepackage{amsthm}

\usepackage{xcolor,colortbl}
\definecolor{Gray}{gray}{0.90}
\newcolumntype{g}{>{\columncolor{Gray}}c}
\definecolor{ffe1da}{RGB}{255,225,218}
\definecolor{E6F5F0}{HTML}{E6F5F0}
\definecolor{darkF7E0D5}{RGB}{209,154,128}
\colorlet{Light}{E6F5F0}
\newcommand{\CC}[1]{\cellcolor{Light}}

\usepackage{amssymb}
\usepackage{pifont}

\newcommand{\lavila}{{\textsc{LaViLa}}}
\newcommand{\model}{EgoVLPv2}

\usepackage[pagebackref=true,breaklinks=true,letterpaper=true,colorlinks,bookmarks=false]{hyperref}

\iccvfinalcopy 

\def\iccvPaperID{11180} 

\ificcvfinal\fi

\begin{document}

\title{EgoVLPv2: Egocentric Video-Language Pre-training with \\ Fusion in the Backbone}

\author{Shraman Pramanick$^{1,2 \dagger}$ \ \ \ \  Yale Song$^{2}$  \ \ \ \ Sayan Nag$^{3}$ \ \ \ \  Kevin Qinghong Lin$^{4}$ \ \ \ \  Hardik Shah$^{2}$ \\ Mike Zheng Shou$^{4}$ \ \ \ \ Rama Chellappa$^{1}$  \ \ \ \ Pengchuan Zhang$^{2}$\vspace{1em}\\
$^{1}$Johns Hopkins University, $^{2}$Meta AI, $^{3}$University of Toronto, $^{4}$National University of Singapore \\
}

\maketitle
\ificcvfinal\thispagestyle{empty}\fi



\begin{abstract}

Video-language pre-training (VLP) has become increasingly important due to its ability to generalize to various vision and language tasks. However, existing egocentric VLP frameworks utilize separate video and language encoders and learn task-specific cross-modal information only during fine-tuning, limiting the development of a unified system. In this work, we introduce the second generation of egocentric video-language pre-training (\model), a significant improvement from the previous generation, by incorporating cross-modal fusion directly into the video and language backbones. \model\ learns strong video-text representation during pre-training and reuses the cross-modal attention modules to support different downstream tasks in a flexible and efficient manner, reducing fine-tuning costs. Moreover, our proposed fusion in the backbone strategy is more lightweight and compute-efficient than stacking additional fusion-specific layers. Extensive experiments on a wide range of VL tasks demonstrate the effectiveness of \model\ by achieving consistent state-of-the-art performance over strong baselines across all downstream. Our project page can be found at \href{https://shramanpramanick.github.io/EgoVLPv2/}{https://shramanpramanick.github.io/EgoVLPv2/}.
\blfootnote{$^\dagger$Part of this work was done during an internship at Meta AI.}
 
\end{abstract}

\vspace{-2mm}
\section{Introduction}

Video-Language Pre-training (VLP) has proven to be the \textit{de-facto} solution for a variety of video-text tasks, e.g., video-text retrieval \cite{xu2016msr, patricksupport, bain2021frozen}, VQA \cite{xu2017video, yu2018joint, zhu2020actbert}, zero-shot recognition, \cite{brattoli2020rethinking, lin2022cross, kerrigan2021reformulating} and video-text grounding \cite{mun2020local, lin2023univtg}. This is fueled by recent advances in vision \cite{dosovitskiy2021an, liu2021swin, bertasius2021space, bain2021frozen, arnab2021vivit, fan2021multiscale, liu2022video} and language \cite{vaswani2017attention, devlin-etal-2019-bert, liu2019roberta, yang2019xlnet, sanh2019distilbert, conneau2019cross, raffel2020exploring}, coupled with large-scale data \cite{xu2016msr, zhou2018towards, miech2019howto100m, bain2021frozen, grauman2022ego4d, damen2022rescaling}. Existing video-language datasets generally fall under two categories: third-person view and first-person view (egocentric). The noticeable domain gap between them restricts VLP frameworks pre-trained on third-person videos from performing well on egocentric benchmarks \cite{linegocentric}. However, the recent introduction of a massive-scale egocentric dataset Ego4D \cite{grauman2022ego4d} helps unlock the full potential of egocentric VLP.




\begin{figure}[!t]
\vspace{8pt}
\hspace{-25pt}
\resizebox{1.1\linewidth}{!}{
    \newcommand{\lattice}{4}
\newcommand{\naxis}{8}
\newcommand{\amax}{61.9}
\newcommand{\amin}{59.4}

\newcommand{\bmax}{47.3}
\newcommand{\bmin}{45.0}

\newcommand{\cmax}{52.1}
\newcommand{\cmin}{49.7}

\newcommand{\dmax}{60.9}
\newcommand{\dmin}{57.2}

\newcommand{\emax}{23.8}
\newcommand{\emin}{18.8}

\newcommand{\fmax}{34.1}
\newcommand{\fmin}{32.1}

\newcommand{\gmax}{68.2}
\newcommand{\gmin}{65.6}

\newcommand{\hmax}{37.9}
\newcommand{\hmin}{32.7}

\newcommand{\origin}{0.7} 

\newcommand\ColorBox[1]{\textcolor{#1}{\rule{3ex}{3ex}}}

\newcommand{\annotMark}[5]{
	\pgfmathsetmacro{\xcor}{#3*cos{(#1*#2)}/(1/#4)};
	\pgfmathsetmacro{\ycor}{#3*sin{(#1*#2)}/(1/#4)};
	\draw (\xcor,\ycor)node[anchor=south]{\large #5};
}

\begin{tikzpicture}[rotate=0, scale=0.95,every node/.style={inner sep=-15,outer sep=-15}]
	\tkzKiviatDiagram[lattice=\lattice, gap=1, step=1, label space=1.5]
	{EK-100\\ MIR \\(nDCG),
		EK-100\\ MIR\\ (mAP),
		QFVS \\(avg F-score),
		EgoMCQ \\(intra-vid. acc.),
		EgoNLQ (R@5 IoU@0.3),
		CharadesEgo (mAP),
		EgoMQ (R@5 IoU@0.3),
		EgoTaskQA \\(acc.)}
	
		\tkzKiviatLine[thick, fill=color_our!80, color=NiceGreen, opacity=0.5](
		\fpeval{(61.9/\amax-\origin)/(1 - \origin)*\lattice},
		\fpeval{(47.3/\bmax-\origin)/(1 - \origin)*\lattice},
		\fpeval{(52.1/\cmax-\origin)/(1 - \origin)*\lattice},
		\fpeval{(60.9/\dmax-\origin)/(1 - \origin)*\lattice},
		\fpeval{(23.8/\emax-\origin)/(1 - \origin)*\lattice},
		\fpeval{(34.1/\fmax-\origin)/(1 - \origin)*\lattice},
		\fpeval{(68.2/\gmax-\origin)/(1 - \origin)*\lattice},
		\fpeval{(37.9/\hmax-\origin)/(1 - \origin)*\lattice})
		\tkzKiviatLine[thick, fill=gray!50, color=NiceGray, opacity=0.8](
		\fpeval{(59.4/\amax-\origin)/(1 - \origin)*\lattice},
		\fpeval{(45.0/\bmax-\origin)/(1 - \origin)*\lattice},
		\fpeval{(49.7/\cmax-\origin)/(1 - \origin)*\lattice},
		\fpeval{(57.2/\dmax-\origin)/(1 - \origin)*\lattice},
		\fpeval{(18.8/\emax-\origin)/(1 - \origin)*\lattice},
		\fpeval{(32.1/\fmax-\origin)/(1 - \origin)*\lattice},
		\fpeval{(65.6/\gmax-\origin)/(1 - \origin)*\lattice},
		\fpeval{(32.7/\hmax-\origin)/(1 - \origin)*\lattice})
	\annotMark{0.15}{360/\naxis}{2.4}{1}{\amin};
	\annotMark{0.09}{360/\naxis}{4.35}{1}{\amax};
	\annotMark{1}{360/\naxis}{2.9}{1}{\bmin};
	\annotMark{1}{360/\naxis}{4.7}{1}{\bmax};
	\annotMark{2}{325/\naxis}{3.2}{1}{\cmin};
	\annotMark{2}{335/\naxis}{4.8}{1}{\cmax};
	\annotMark{3}{310/\naxis}{3.0}{1}{\dmin};
	\annotMark{3}{330/\naxis}{4.6}{1}{\dmax};
	\annotMark{4}{313/\naxis}{1.98}{1}{\emin};
	\annotMark{4}{333/\naxis}{3.4}{1}{\emax};
	\annotMark{5}{337/\naxis}{3.1}{1}{\fmin};
	\annotMark{5}{340/\naxis}{4.0}{1}{\fmax};
	\annotMark{6}{343/\naxis}{2.3}{1}{\gmin};
	\annotMark{6}{349/\naxis}{3.7}{1}{\gmax};
	\annotMark{7}{358/\naxis}{1.4}{1}{\hmin};
	\annotMark{7}{358/\naxis}{3.0}{1}{\hmax};
	\node[anchor=south west,xshift=-34pt,yshift=20pt] at (current bounding box.south east)
{
	\begin{tabular}{@{}lp{3cm}@{}}
		\ColorBox{color_our!80} & \large \model\\
		\ColorBox{gray!50} & \large EgoVLP~\cite{linegocentric} \\
	\end{tabular}
};
\end{tikzpicture}%
}
\vspace{.5pt}
\caption{\textbf{\model\ achieves the state-of-the-art} performance across a broad range of egocentric video understanding tasks (see Table \ref{tab:downstream_tasks} for details) among similar-sized baselines by incorporating cross-modal attention in the transformer backbones to learn video-language representation.}   

\label{fig:results_summary}
\vspace{-2mm}
\end{figure}
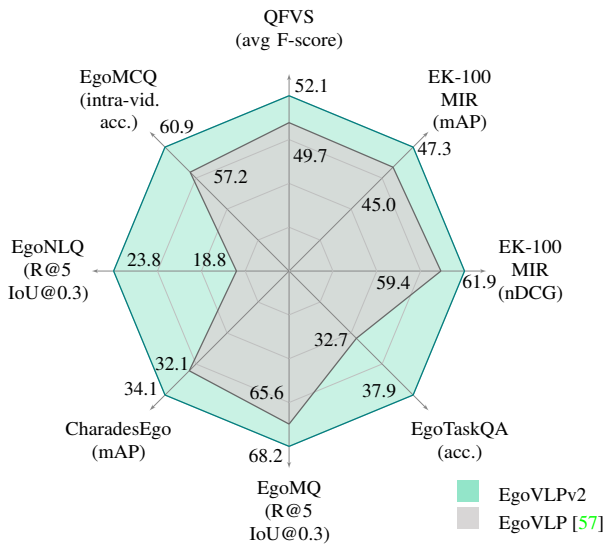

\begin{figure*}[!t]
 \centering
 \begin{subfigure}[b]{0.24\textwidth}
     \centering
     \includegraphics[width=\textwidth]{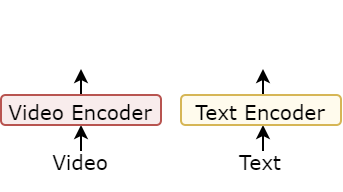}
     \caption{Dual Encoders}
     \label{fig:}
 \end{subfigure}
 \hfill
 \begin{subfigure}[b]{0.24\textwidth}
     \centering
     \includegraphics[width=\textwidth]{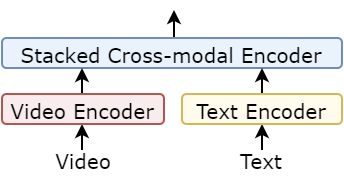}
     \caption{Stacked Fusion Layers}
     \label{fig:}
 \end{subfigure}
 \hfill
 \begin{subfigure}[b]{0.24\textwidth}
     \centering
     \includegraphics[width=\textwidth]{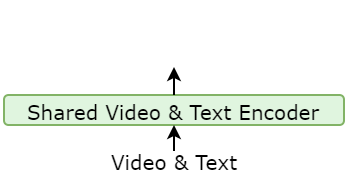}
     \caption{Shared Encoders}
     \label{fig:}
 \end{subfigure}
 \begin{subfigure}[b]{0.24\textwidth}
     \centering
     \includegraphics[width=\textwidth]{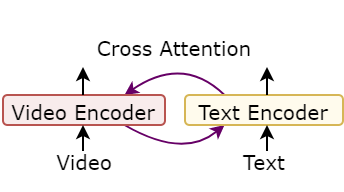}
     \caption{Fusion in the Backbone (Ours)}
     \label{fig:}
 \end{subfigure}
\vspace{-.5mm}
     \caption{\textbf{Four categories of VLP frameworks.} 
(a) use separate (\textit{dual}) video and text backbones, with InfoNCE~\cite{oord2018representation} as the common pretraining objective \cite{linegocentric, zhao2022learning, ashutosh2023hiervl, moon2022imu2clip}
(b) use cross-modal fusion layers on top of dual encoders, with MLM, VTM, etc. as common pretraining tasks \cite{luo2020univl, lei2021less, xu2021vlm, sunlong}
(c) use a single encoder for different modalities, with similar learning objectives as (b) \cite{li2022lavender, akbari2021vatt, leeparameter}
(d) Fusion in the Backbone (Ours).}
\vspace{-1.5mm}
\label{fig:system_categories}
\vspace{-1mm}
\end{figure*}

Existing egocentric VLP approaches \cite{linegocentric, zhao2022learning, moon2022imu2clip, ashutosh2023hiervl} pre-train separate (\textit{dual}) video and language encoders and learn task-specific cross-modal information only during fine-tuning, limiting the development of unified egocentric VL frameworks. Moreover, they lack strong zero-shot inference ability on multi-modal downstream tasks. This issue is commonly addressed by stacking dedicated fusion layers on top of the dual video and text encoders \cite{luo2020univl, lei2021less, xu2021vlm, sunlong, xue2022advancing, yang2021taco, zellers2021merlot}, or with a shared video-language architecture \cite{li2022lavender, akbari2021vatt, leeparameter, tang2023perceiver, wang2023all}. However, these approaches introduce a large number of fusion-specific parameters, and the resulting encoder cannot be directly applied to uni-modal (video-only) tasks.

In this work, we present the second generation of egocentric VLP (EgoVLPv2), 
a significant improvement over the previous generation~\cite{linegocentric} by incorporating cross-modal fusion directly into the video and language backbones. Our approach improves over existing VLP frameworks by: $(i)$ fewer fusion parameters compared to stacked fusion-specific transformer layers or shared encoders, requiring less GPU memory, compute resources, and training time; $(ii)$ the flexibility to switch between dual and fusion encoders, by turning on and off cross-attention fusion using a gating mechanism; $(iii)$ being applicable to both uni- and multi-modal tasks. 


Inserting cross-modal fusion directly into the backbone helps unify a wide range of dual- and fusion-encoder-based downstream tasks. Specifically, the ``switching'' ability of \model\ enables us to utilize the same pre-trained encoders for fast retrieval and grounding tasks, which require dual and fusion encoders, respectively. Moreover, in contrast to existing egocentric VLP frameworks that learn task-specific fusion parameters during fine-tuning, \model\ reuses the pre-trained cross-attention modules across different tasks, significantly reducing the fine-tuning cost. This enables us to introduce query-focused video summarization as a downstream task, which has recently gained attention in the community \cite{nalla2020watch, wu2022intentvizor, xiao2020query, jiang2019hierarchical, xiao2020convolutional, narasimhan2021clip}. The scarcity of annotated data has been a bottleneck to training decent-sized models end-to-end on this task, with the only available egocentric dataset, QFVS \cite{sharghi2017query}, providing merely $135$ video-query training samples. \model\ achieves new state-of-the-art results on QFVS with a decent margin over the baselines.


In summary, our contributions are: $(i)$ We advance a step forward in egocentric VLP by proposing \model, the second generation of EgoVLP~\cite{linegocentric} with cross-modal fusion in the backbone. Our proposed framework can switch between dual and fusion encoders and requires 45\% lesser compute (GMACs) than learning additional fusion-specific transformer layers. 
$(ii)$ The switching capability of \model\ allows us to unify a wide range of dual- and fusion-encoder-based downstream tasks under the same VLP framework and reduce the task-specific fine-tuning cost by employing the same pre-trained cross-attention modules across different video-language tasks.  
$(iii)$ We demonstrate the effectiveness of \model\ on eight egocentric benchmarks and achieve state-of-the-art performance among comparable-sized backbones. We summarize these results in Figure \ref{fig:results_summary}. 



\section{Related Works}

\subsection{VLP Frameworks}
Video-language pre-training (VLP) has attracted increasing attention in recent years, following the success of image-language pre-training \cite{radford2021learning, li2021align, jia2021scaling, doucoarse, baovlmo, chen2020uniter, lu2019vilbert, li2020oscar, dou2022empirical, zhai2022lit, yang2022unified, yang2022unitab, pramanick2022volta, flip2023, git2023, beit2023, oneR2023, ptp2023, softmask2023, blip22023} and their applications \cite{chen2019neural, fu2015tagging, huang2021unifying, li2019emotion, pramanick2022multimodal}. There are three broad categories of VLP frameworks (see Figure \ref{fig:system_categories}):

\vspace{1mm}

\begin{figure*}[!t]
\centering
\includegraphics[width=0.98\textwidth]{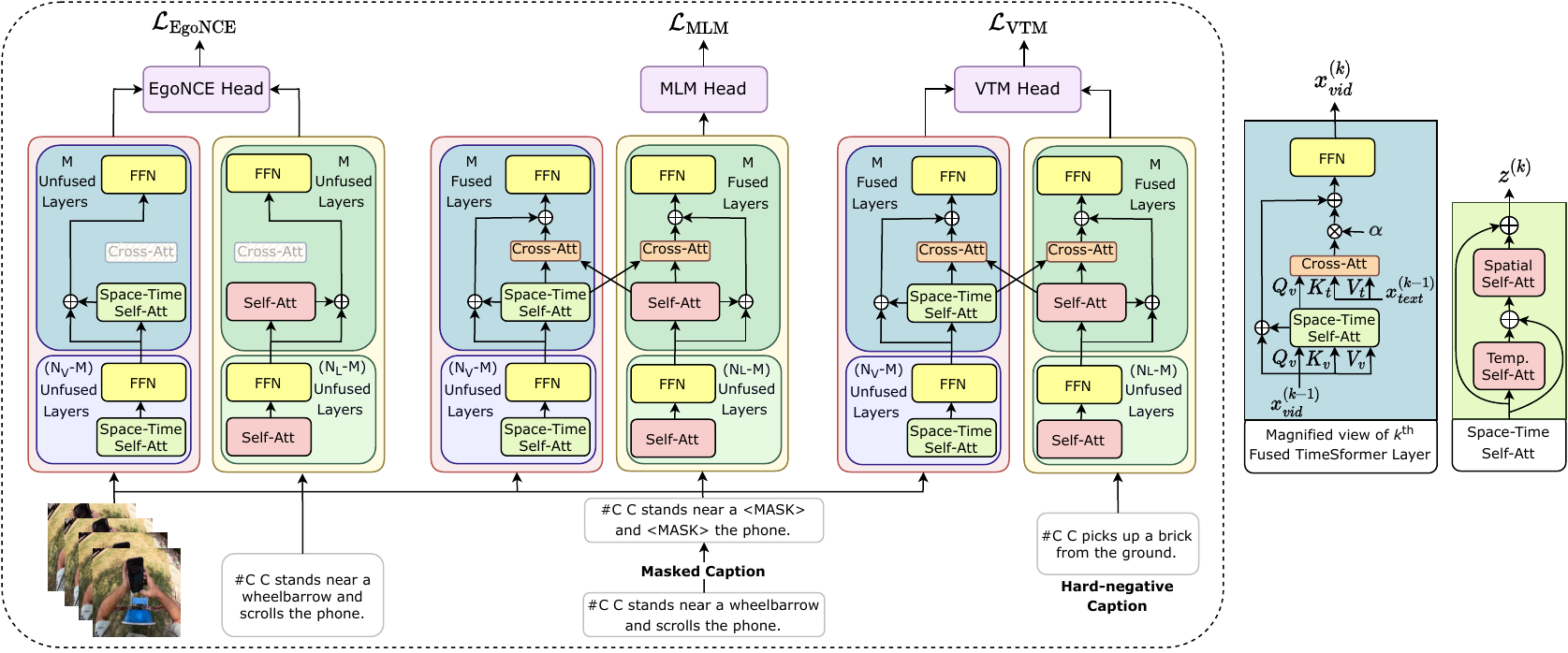}
\caption{\textbf{Computation of three objectives, $\mathcal{L}_\mathrm{EgoNCE}$, $\mathcal{L}_\mathrm{MLM}$, and $\mathcal{L}_\mathrm{VTM}$.} We insert cross-modal fusion into uni-modal backbones with a gating mechanism. During pre-training, every forward iteration contains three steps: $(i)$ cross-attention modules are switched off, \model\ acts as dual encoder, $\mathcal{L}_\mathrm{EgoNCE}$ is computed. $(ii)$ cross-attention is switched on, \model\ acts as fusion encoder, and video-masked narration pair is fed into \model\ to compute $\mathcal{L}_\mathrm{MLM}$ $(iii)$ cross-attention is kept on, hard-negative video-narration pairs are fed into \model\ to compute $\mathcal{L}_\mathrm{VTM}$. This \textit{fusion in the backbone} strategy results in a lightweight and flexible model compared to using fusion-specific transformer layers.}
\label{fig:system_overview}
\vspace{-2mm}
\end{figure*}

\noindent \textbf{Dual Encoders:} Many existing egocentric VLP frameworks \cite{linegocentric, zhao2022learning, ashutosh2023hiervl, moon2022imu2clip} falls into this category. They use separate video and language backbones and learn task-specific cross-modal fusion during fine-tuning~\cite{bain2021frozen, miech2020end, xu2021videoclip, wang2022object}. They are commonly trained using InfoNCE \cite{oord2018representation} or MIL-NCE \cite{miech2020end} objectives, and have been successful in video-text retrieval. 

\vspace{1mm}

\noindent \textbf{Shared Encoder:} Approaches that learn a combined encoder for video and text fall under this category \cite{li2022lavender, akbari2021vatt, leeparameter, tang2023perceiver, wang2023all}. They are modality independent and can be applied to an image, video, text, audio, time-series, and single-view 3D data. Common learning objectives include masked language modeling \cite{devlin-etal-2019-bert, zhu2020actbert}, masked frame modeling \cite{sun2019videobert, zhu2020actbert}, masked token modeling \cite{xu2021vlm}, masked modal modeling \cite{luo2020univl, xu2021vlm}, sentence ordering modeling \cite{lei2021understanding}, frame ordering modeling \cite{lei2021understanding, li2020hero}, and video-text matching \cite{lei2021understanding}.  

\vspace{1mm}

\noindent \textbf{Encoders with Stacked Fusion Layers:} This line of work uses dedicated cross-modal fusion layers on top of dual encoders~\cite{luo2020univl, lei2021less, xu2021vlm, sunlong, xue2022advancing, yang2021taco, zellers2021merlot}, trained using similar objectives as shared encoders. 

The latter two categories introduce a large number parameters for cross-modal fusion. In this work, we propose a fourth category (Figure \ref{fig:system_categories} (d)) by inserting cross-modal fusion in uni-modal backbones using a gating mechanism. Our framework is flexible to act as either dual or shared encoders by switching cross-attention modules off and on. 

\subsection{Video-Language Datasets}
The success of VLP can be partially attributed to the availability of large-scale open-world video-text datasets such as ActivityNet \cite{krishna2017dense}, WebVid-$2$M \cite{bain2021frozen}, and HowTo$100$M \cite{miech2019howto100m}.
These datasets comprise videos sourced from the Web, and are paired with the corresponding ASR captions, making them popular for VLP pre-training. 
Despite their impressive size, these existing video-text pretraining datasets typically feature 3rd-person views.
On the other hand, egocentric videos has received increasing interests from the community. Previous egocentric datasets \cite{damen2022rescaling, sigurdsson2018charades, li2015delving, shah2023steps, plizzari2023outlook} were small-scale and domain-specific. The recently released Ego4D \cite{grauman2022ego4d} is the first massive-scale egocentric dataset consisting of $3670$ hours of videos
collected by 931 people from 74 locations across 9 different countries world-wide. Recently, EgoClip \cite{linegocentric} offered a filtered version of Ego4D with variable-length clip intervals instead of single timestamps. We train our proposed framework, \model, on the EgoClip version of Ego4D. 


\section{\model}

\subsection{Fusion in the Backbone} \label{sec:fusion_in_the_backbone}

We use TimeSformer \cite{bertasius2021space, bain2021frozen} and RoBERTa \cite{liu2019roberta} as our video and language backbones. However, such separate (\textit{dual}) uni-modal encoder design does not capture cross-modality interaction and, thus, fails to produce fine-grained multi-modal representation. Existing VLP frameworks achieve cross-modal fusion by: $(i)$ learning a shared architecture \cite{li2022lavender, akbari2021vatt, leeparameter, tang2023perceiver, wang2023all} or stack fusion layers on top of dual encoders \cite{luo2020univl, lei2021less, xu2021vlm, sunlong, xue2022advancing, yang2021taco, zellers2021merlot}, or $(ii)$ learning cross-modal fusion during fine-tuning \cite{linegocentric, zhao2022learning, ashutosh2023hiervl, moon2022imu2clip, bain2021frozen, miech2020end, xu2021videoclip, wang2022object}. While the former offers superior cross-modal representation and zero-shot inference ability on multi-modal downstream tasks, they introduce a large number of fusion parameters than the latter. In this work, we insert cross-modal fusion into the top few layers of uni-modal backbones to strike a balance between the two ideas. 


Figure \ref{fig:system_overview} shows the architecture of \model. Each TimeSformer encoder layer has a divided space-time attention module containing temporal and spatial self-attentions with residual connections. The output of space-time attention at $k^{th}$ encoder layer, $z^{(k)}$, can be expressed as:
\begin{align}\label{eq:space-time}
\vspace{-5mm}
     \hat{x}^{(k)}_{vid} &= x^{(k-1)}_{vid} + \textsc{Temp-SA}(x^{(k-1)}_{vid}) \nonumber \\
     z^{(k)} &= x^{(k-1)}_{vid} + \textsc{Spa-SA}(\hat{x}^{(k)}_{vid}) \nonumber \\ 
     &= \textsc{Space-Time}(x^{(k-1)}_{vid})
\end{align}
where $x^{(k-1)}_{vid}$ is the output of the $(k-1)^{th}$ encoder layer, $\textsc{Temp-SA}$ and $\textsc{Spa-SA}$ represent temporal and spatial self-attention blocks, respectively. We insert multi-modal fusion inside the backbone by introducing gated cross-attention after the space-time attention module. Hence, the output of $k^{th}$ fused TimeSformer layer, $x^{(k)}_{vid}$\,, can be expressed as: 
\begin{align}\label{eq:cross_attention_video}
\vspace{-3mm}
    z^{(k)} &= \textsc{Space-Time}(x^{(k-1)}_{vid}) \nonumber \\
    x^{(k)}_{vid} &= x^{(k-1)}_{vid} +  z^{(k)} + \alpha * \textsc{CA}( z^{(k)}, x^{(k-1)}_{text})\\
    x^{(k)}_{vid} &=  x^{(k)}_{vid} + \textsc{FFN}(x^{(k)}_{vid}) \nonumber
\end{align}
where $x^{(k-1)}_{text}$ is the output from the $(k-1)^{th}$ RoBERTa layer, $\textsc{CA}$, $\textsc{FFN}$ denote cross-attention block and feed-forward network, respectively, and $\alpha$ is a learnable gating parameter initialized from $0$. Each RoBERTa layer contains multi-head self-attention~\cite{vaswani2017attention} followed by feed-forward layers. Similar to the fused TimeSformer module, we insert cross-attention into the RoBERTa backbone:
\begin{align}\label{eq:cross_attention_text}
\vspace{-3mm}
    \hat{x}^{(k)}_{text} &= \textsc{SA}(x^{(k-1)}_{text}) \nonumber \\
    x^{(k)}_{text} &= x^{(k-1)}_{text} + \hat{x}^{(k)}_{text} + \alpha * \textsc{CA}(\hat{x}^{(k)}_{text},  x^{(k)}_{vid})\\
    x^{(k)}_{text} &= x^{(k)}_{text} + \textsc{FFN}(x^{(k)}_{text}) \nonumber
\end{align}
where $\textsc{SA}$ is the traditional self-attention module. For simplicity, we insert cross-attention into the same number of layers in both backbones. Notably, such \textit{fusion in the backbone} strategy is not only limited to TimeSformer and RoBERTa; but can also be applied to any transformer-based video~\cite{liu2022video, fan2021multiscale, arnab2021vivit} and text~\cite{devlin-etal-2019-bert, sanh2019distilbert, yang2019xlnet} encoders.

Fusion in the backbone with gated cross-attention has the following advantages: $(i)$ Cross-attention parameters can easily
be switched off by setting the gating scalar $\alpha$ to 0; thus, the model behaves as a dual encoder, which is helpful for scenarios that require ``unfused'' video and textual features; $(ii)$ Our fusion approach is more lightweight and compute-efficient than adding fusion-specific transformer layers, which is demonstrated in detail in Section \ref{sec:ablation_study}. 


\subsection{Pre-training Objectives} \label{sec:pre_training_objectives}
We use three pre-training objectives: $(1)$ Egocentric noise contrastive estimation (EgoNCE), $(2)$ masked language modeling (MLM), and $(3)$ video-text matching (VTM). 

\vspace{1mm}


\noindent \textbf{EgoNCE:} Lin et al.~\cite{linegocentric} proposed EgoNCE for dual-encoder-based egocentric VLP. It makes two modifications over InfoNCE \cite{oord2018representation}: $(i)$ Besides the matched video-text samples, all pairs that share at least one noun or one verb are treated as positives. $(ii)$ Every batch of $N$ video-text samples is augmented with another $N$ visually similar videos, which are treated as additional negatives. Overall, video-to-text EgoNCE objective, $\mathcal{L}^\mathrm{ego}_\mathrm{v2t}$, can be expressed as:
\begin{align}\label{eq:egonce_video_to_text}
\vspace{-5mm}
	\mathcal{L}^\mathrm{ego}_\mathrm{v2t}=\frac{1}{| \mathcal{\widetilde{B}} |}\sum_{i\in\mathcal{\widetilde{B}}}  \log 
	\frac{
	\textcolor{brown}{
	\sum\limits_{k\in \mathcal{P}_\mathrm{i}}\exp\left(\frac{\mathbf{v}_\mathrm{i}^T\mathbf{t}_k }{\tau}\right)
	}
	}
	{  \sum\limits_{j\in \mathcal{B}} \left( \exp\left(\frac{\mathbf{v}_\mathrm{i}^T\mathbf{t}_\mathrm{j} }{\tau}\right) +
	\textcolor{blue}{\exp\left(\frac{\mathbf{v}_\mathrm{i}^T\mathbf{t}_\mathrm{j'}} {\tau}\right)}  \right) 
	}
\end{align}
where the $i^{th}$ video embedding $v_i$ and $j^{th}$ text embedding $t_j$ are L$_2$ normalized features, and $\tau$ is a temperature factor. $\widetilde{B}$ is the augmented batch with $2N$ samples. The term in \textcolor{brown}{brown} are the modified positive samples, and the term in \textcolor{blue}{blue} are the modified negative samples. The text-to-video EgoNCE objective, $\mathcal{L}^\mathrm{ego}_\mathrm{t2v}$, can be defined symmetrically. The total EgoNCE loss is: $\mathcal{L}_\mathrm{EgoNCE} = \mathcal{L}^\mathrm{ego}_\mathrm{v2t} + \mathcal{L}^\mathrm{ego}_\mathrm{t2v}$. 

We compute EgoNCE in a dual-encoder setting. Specifically, we set $\alpha = 0$, and thus, the cross-attention modules are switched off to calculate the EgoNCE loss. 


\begin{figure}
 \centering
 \begin{subfigure}{0.1496\textwidth}
     \centering
     \includegraphics[width=\textwidth]{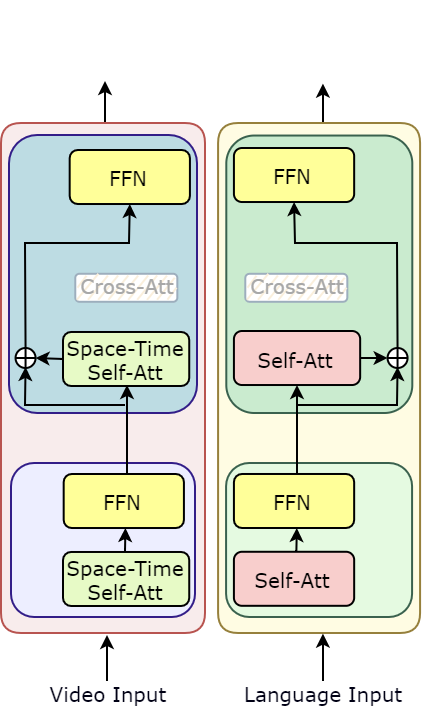}
     \caption{Retrieval w/ Dual Encoder.}
     \label{fig:retrieval}
 \end{subfigure}
 \hfill
 \begin{subfigure}{0.155\textwidth}
     \centering
     \includegraphics[width=\textwidth]{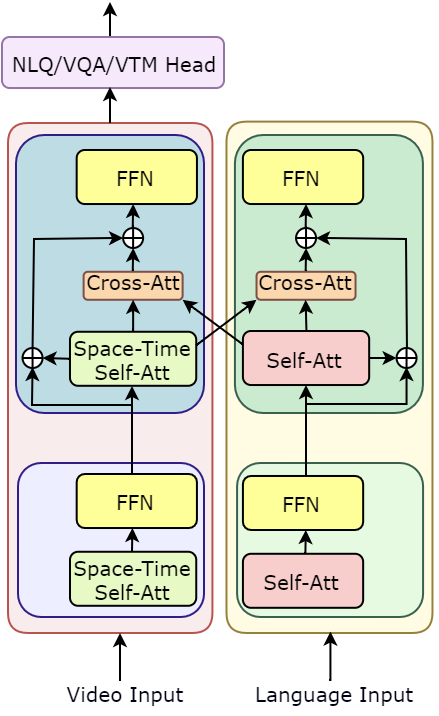}
     \caption{VQA/retrieval w/ Fusion Encoder.}
     \label{fig:nlq}
 \end{subfigure}
 \hfill
 \begin{subfigure}{0.1523\textwidth}
     \centering
     \includegraphics[width=\textwidth]{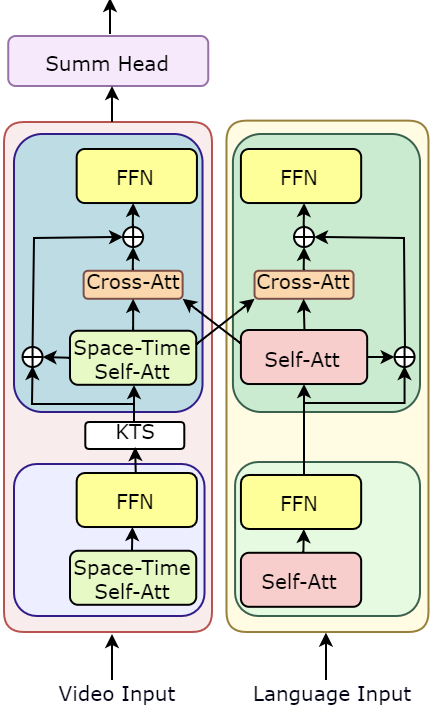}
     \caption{QFVS w/ Fusion Encoder.}
     \label{fig:summ}
 \end{subfigure}
\caption{\textbf{\model\ can be adapted to various dual- and fusion-encoder-based video-language tasks,} ranging from retrieval, video question-answering, and video grounding to query-focused video summarization.}
\label{fig:downstream_adaptation}
\vspace{-2mm}
\end{figure}

\vspace{1mm}

\noindent \textbf{MLM:} Masked language modeling and video-text matching are proven helpful in fusion-encoder-based VLP literature \cite{devlin-etal-2019-bert, zhu2020actbert}. For MLM, we randomly mask $15\%$ text tokens,\footnote{Following BERT, we decompose this $15\%$ into $10\%$ random words, $10\%$ unchanged, and $80\%$ with a special token [MASK].
} and the loss, $\mathcal{L}_\mathrm{MLM}$, aims to reconstruct the masked tokens based on surrounding words and video patches by minimizing the negative log-likelihood. 


\vspace{1mm}

\noindent \textbf{VTM:} For the VTM objective, the model is given a video-text sample, and the output is a binary label $ y\in \{0,1\}$ indicating if the input pair is matched. $\mathcal{L}_\mathrm{VTM}$ is constructed as a binary cross-entropy loss over the predicted scores. Following \cite{baovlmo, doucoarse}, we sample the global hard-negative video-text pairs using the similarities computed by EgoNCE. 

We compute $\mathcal{L}_\mathrm{MLM}$ and $\mathcal{L}_\mathrm{VTM}$ in a fusion-encoder setting. In this case, $\alpha \neq 0$ and the cross-attention modules are switched on. Overall, our \model\ pre-training pipeline can be summarized in the following three steps: 


\begin{itemize}[leftmargin=*]
\vspace{-2mm}
    \item \textbf{EgoNCE} requires unfused video and text features, so we switch off cross-attention ($\alpha = 0$). Thus, $\mathcal{L}_\mathrm{EgoNCE}$ is computed with \model\ acting as a dual encoder. 
    \item \textbf{MLM \& VTM} requires multi-modal representation. We switch on cross-attention modules and compute $\mathcal{L}_\mathrm{MLM}$ and $\mathcal{L}_\mathrm{VTM}$ with \model\ acting as a fusion encoder.
    \item For \textbf{back-propagation}, the three losses are added, resulting in $\mathcal{L}_\mathrm{total} = (1 - \gamma - \delta) \mathcal{L}_\mathrm{EgoNCE} + \gamma \mathcal{L}_\mathrm{MLM} + \delta \mathcal{L}_\mathrm{VTM}$, and back-propagated into the model end-to-end. $\gamma$ and $\delta$ are hyper-parameters that control the contribution of different terms on $\mathcal{L}_\mathrm{total}$. An ablation on different pre-training objectives of \model\ is provided in Section \ref{sec:ablation_study}. The pseudo-code for pre-training \model\ can be found in the supplementary. 
    
\end{itemize}


\subsection{Adaptation to Downstream Tasks} 

We now describe how we adapt \model\ to different downstream tasks as shown in Figure \ref{fig:downstream_adaptation}. 

\vspace{1mm}

\noindent \textbf{Video-Text Retrieval:} We perform retrieval in two settings: $(i)$ \textit{dual encoders:} we switch off cross-attention and use \model\ as a dual encoder, and compute the cosine similarity between video clips and text narrations. $(ii)$ \textit{fusion encoders:} we switch on cross-attention. The top $M$ layers of the video and language backbones interact and produce multi-modal representations, which are fed into the pre-trained VTM head to compute matching scores. We also compute an ensemble of the two to further boost the performance, discussed in Section \ref{sec:ablation_study}. 

\vspace{1mm}

\noindent \textbf{Video Grounding and Question Answering:} We perform both uni- (video-only) and multi-modal (text-guided) video grounding. We switch off cross-attention for uni-modal grounding and use only the video encoder. We use \model\ as a fusion encoder for text-guided grounding and video question answering.  

\vspace{1mm}

\noindent \textbf{Query-focused Video Summarization:} The input videos are very long ($3$-$5$ hours) for this task. We first use the unfused $N-M$ layers\footnote{For simplicity, we keep the number of unfused and fused layers the same in the video and text encoder.} of our video and text encoders to extract uni-modal features from 5 second clips and the text query. Next, we apply the KTS shot boundary detector \cite{potapov2014category} to segment the long video.
After this, the query and segment-wise clip features are fed into the top $M$ fused layers of \model\ to compute the multi-modal representation. Finally, we learn an additional single-layer transformer to design the interrelation across all $5$ second long clips in every segment. We present additional details for the query-focused video summarization framework in the supplementary. 

\noindent \textbf{}

\label{sec:adaptation_downstreams}

\begin{table}[!t]
\centering

\small
\setlength{\tabcolsep}{4pt}
\resizebox{\columnwidth}{!}{\begin{tabular}{@{}l c| c c | c | c @{}}

\toprule
\multirow{2}{*}{\textbf{Dataset}} & \multicolumn{1}{c|}{\multirow{2}{*}{\textbf{Task}}} & \multirow{2}{1.0cm}{\textbf{Multi-modal}} & \multirow{2}{*}{\textbf{Fusion}} & \multirow{2}{*}{\textbf{Metrics (\%)}} & \multirow{2}{*}{\textbf{Eval.}}  \\ 

& & & & & \\

\midrule 

\multirow{4}{*}{Ego4D \cite{grauman2022ego4d}} & \multirow{1}{*}{MCQ w/ dual} & \ding{51} & \ding{55} & Inter- \& Intra Acc. & ZS \\
& MCQ w/ fusion & \ding{51} & \ding{51} & Inter- \& Intra Acc.  & ZS \\

& NLQ & \ding{51} & \ding{51} & Recall\,$@$\,N & HT \\
& MQ & \ding{55} & $-$ & mAP, Recall\,$@$\,N & HT \\

\midrule

QFVS \cite{sharghi2017query} & Video Summ. & \ding{51} & \ding{51} & F-$1$ & HT\\
EgoTaskQA \cite{jiaegotaskqa} & Video QA & \ding{51} & \ding{51} & Mean Acc. & HT, FT \\
\multirow{1}{*}{CharadesEgo \cite{sigurdsson2018charades}} & CLS$^\dagger$ & \ding{51} & \ding{55} & Video-level mAP & ZS, FT \\
\multirow{1}{*}{EK-100 \cite{damen2022rescaling}} & MIR w/ dual & \ding{51} & \ding{55} & mAP, nDCG & ZS, FT\\

\bottomrule
\end{tabular}}
\caption{\textbf{Egocentric downstream datasets, metrics, and evaluation protocols.} We evaluate \model\ on a wide variety of benchmarks: video-text retrieval (EgoMCQ, CharadesEgo, EK-100), uni-modal and text-guided video grounding (EgoMQ, EgoNLQ), video question answering (EgoTaskQA) and query-focused video summarization (QFVS). The evaluation protocols include zero-shot (ZS), task-specific head-tuning (HT), and end-to-end fine-tuning (FT). $^\dagger$ChardesEgo is a multi-class classification problem, but we convert this to a retrieval task. Please find more details in Section \ref{sec:pretraining_downstream_datasets} and in supplementary.}\label{tab:downstream_tasks}
\vspace{-2mm}
\end{table}

\section{Experiments}

\subsection{Pre-training \& Downstream Datasets} \label{sec:pretraining_downstream_datasets}
We pre-train \model\ on the EgoClip \cite{linegocentric} version of Ego4D \cite{grauman2022ego4d}, the largest publicly available egocentric video dataset. EgoClip sources untrimmed egocentric videos from Ego4D and offers filtered video-narration samples with variable-length clip intervals instead of single timestamps of Ego4D. Moreover, EgoClip excludes the videos appearing in the validation and test sets of the Ego4D benchmark \cite{grauman2022ego4d}, resulting in around $3.8$M pre-training samples covering over $2927$ hours of video from $129$ different scenarios. 

We evaluate \model\ across multiple benchmarks on five egocentric datasets, summarized in Table \ref{tab:downstream_tasks}: 
\begin{itemize}[leftmargin=*]
\vspace{-2mm}

\item On Ego4D \cite{grauman2022ego4d} benchmarks: Multiple-Choice Questions (\textbf{EgoMCQ}) is a text-to-video (T $\to$ V) retrieval task with five video clips for every query text. Natural Language Query (\textbf{EgoNLQ}) is a natural language grounding \cite{hendricks2018localizing, gao2017tall, soldan2021vlg} task that aims to localize a single time interval within a video given a text query. Moment Query (\textbf{EgoMQ}) is a video-only temporal action localization \cite{caba2015activitynet} task. 

\item Query-focused video summarization (\textbf{QFVS}) \cite{sharghi2017query} aims to generate a concise version of a long (3-5 hours) egocentric video based on a natural language query. 

\item Video question-answering on \textbf{EgoTaskQA} \cite{jiaegotaskqa} provides four question types (descriptive, predictive, explanatory, and counterfactual) with direct and indirect references, and evaluates the prediction over spatial, temporal, and causal domains of goal-oriented task understanding. Notably, to the best of our knowledge, we are the first to unify QFVS and EgoTaskQA as two downstream tasks of a VLP framework. 

\item Action Recognition on \textbf{CharadesEgo} \cite{sigurdsson2018charades}: a multi-class classification of daily indoor activities, with class names being short natural language phrases like `\textit{Putting something on a shelf}.' Hence, leveraging text representations with class names, we treat this task as a retrieval problem. 

\item Multi-instance retrieval on Epic-Kitchens-100 \cite{damen2022rescaling} (\textbf{EK-100 MIR}): this is a text-to-video (T $\to$ V) and video-to-text (V $\to$ T) retrieval task, with a significant semantic overlap between different narrations. Detailed statistics of pre-training and downstream datasets and evaluation metrics are given in the supplementary. 
\end{itemize}

\begin{table}[!t]
\centering

\small
\setlength{\tabcolsep}{4pt}
\resizebox{0.99\columnwidth}{!}{\begin{tabular}{@{}l |c | c c | c c c c @{}}
\toprule

\multirow{3}{*}{\bf Method} & \multirow{3}{1.5cm}{\bf \centering \# Pre-train Dataset} & \multicolumn{2}{c |}{\bf EgoMCQ} & \multicolumn{4}{c}{\bf EgoNLQ validation set} \\ 

\cmidrule{3-8}

& & \multicolumn{2}{c |}{Accuracy (\%)} & \multicolumn{2}{c}{mIOU@0.3} & \multicolumn{2}{c}{mIOU@0.5} \\ 

& & Inter & Intra & R@1 & R@5 & R@1 & R@5 \\

\midrule

SlowFast \cite{feichtenhofer2019slowfast} & $-$ & $-$ & $-$ & 5.45 & 10.74 & 3.12 & 6.63 \\
EgoVLP \cite{linegocentric} & 3.8M & \underline{90.6} & \underline{57.2} & \underline{10.84} & \underline{18.84} & \underline{6.81} & \underline{13.45} \\
HierVL-Avg \cite{ashutosh2023hiervl} & 3.8M & 90.3 & 53.1 & $-$ & $-$ & $-$ & $-$ \\
HierVL-SA \cite{ashutosh2023hiervl} & 3.8M & 90.5 & 52.4 & $-$ & $-$ & $-$ & $-$ \\
\demph{\textsc{LaViLa}-B \cite{zhao2022learning}} & \demph{56M} & \demph{93.8} & \demph{59.9} & \demph{10.53} & \demph{19.13} & \demph{6.69} & \demph{13.68}\\
\rowcolor{Light}
\model\ & 3.8M & \bf 91.0 & \bf 60.9 & \bf 12.95 & \bf 23.80 & \bf 7.91 & \bf 16.11 \\

\midrule

 \bf \textcolor{blue}{$\Delta_{\text{Ours - EgoVLP}}$} & $-$ & \textcolor{blue}{0.4} \textcolor{blue}{$\uparrow$} & \textcolor{blue}{3.7} \textcolor{blue}{$\uparrow$} & \textcolor{blue}{2.11} \textcolor{blue}{$\uparrow$} & \textcolor{blue}{4.96} \textcolor{blue}{$\uparrow$} & \textcolor{blue}{1.10} \textcolor{blue}{$\uparrow$} & \textcolor{blue}{2.66} \textcolor{blue}{$\uparrow$} \\

\bottomrule
\end{tabular}}
\caption{\textbf{Performance on EgoMCQ and EgoNLQ's validation set.} \model\ yields significant gains over existing baselines on both tasks. \lavila\ is pre-trained on $15\times$ more narrations generated by GPT-2 \cite{radford2019language}, and is colored \demph{gray}. On EgoMCQ, reported results are achieved by directly ensembling dual- and fusion-encoder-based inference.}\label{tab:egomcq_nlq}
\end{table}

\begin{table}[!t]
\centering

\small
\setlength{\tabcolsep}{4pt}
\resizebox{0.99\columnwidth}{!}{\begin{tabular}{@{}l | c c c c c c | c c c c @{}}
\toprule

\multirow{2}{*}{\bf Method} & \multicolumn{2}{c}{\bf IoU=0.3} & \multicolumn{2}{c}{\bf IoU=0.5} & \multicolumn{2}{c |}{\bf IoU=0.7} & \multicolumn{4}{c}{\bf mAP (\%) @ IoU} \\ 


& R@1 & R@5 & R@1 & R@5 & R@1 & R@5 & 0.1 & 0.3 & 0.5 & Avg. \\

\midrule

SlowFast \cite{feichtenhofer2019slowfast} & 33.45 & 58.43 & 25.16 & 46.18 & 15.36 & 25.81 & 9.10 & 5.76 & 3.41 & 6.03 \\
Frozen \cite{bain2021frozen} & 40.06 & 63.71 & 29.59 & 48.32 & 17.41 & 26.33 & 15.90 & 10.54 & 6.19 & 10.69\\
EgoVLP \cite{linegocentric} & \underline{40.43} & \underline{65.67} & \underline{30.14} & \underline{51.98} & \underline{19.06} & \underline{29.77} & \underline{16.63} & \underline{11.45} & \underline{6.57} & \underline{11.39} \\
\rowcolor{Light}
\model\ & \bf 41.97 & \bf 68.24 & \bf 31.08 & \bf 54.15 & \bf 20.96 & \bf 31.10 & \bf 17.58 & \bf 11.92 & \bf 6.90 & \bf 12.23 \\

\midrule

\bf \textcolor{blue}{$\Delta_{\text{Ours - EgoVLP}}$} & \textcolor{blue}{1.54} \textcolor{blue}{$\uparrow$} & \textcolor{blue}{2.57} \textcolor{blue}{$\uparrow$} & \textcolor{blue}{0.94} \textcolor{blue}{$\uparrow$} & \textcolor{blue}{2.17} \textcolor{blue}{$\uparrow$} & \textcolor{blue}{1.90} \textcolor{blue}{$\uparrow$} & \textcolor{blue}{1.33} \textcolor{blue}{$\uparrow$} & \textcolor{blue}{0.95} \textcolor{blue}{$\uparrow$} & \textcolor{blue}{0.47} \textcolor{blue}{$\uparrow$} & \textcolor{blue}{0.33} \textcolor{blue}{$\uparrow$} & \textcolor{blue}{0.84} \textcolor{blue}{$\uparrow$}\\


\bottomrule
\end{tabular}}
\caption{\textbf{Performance on EgoMQ's validation set.} \model\ sets a new state-of-the-art across all baselines using VSGN \cite{zhao2021video} as grounding head.}\label{tab:egomq}
\vspace{-2mm}
\end{table}

\subsection{Evaluation Protocol}
\noindent We evaluate \model\ using three evaluation protocols:
\begin{itemize}[leftmargin=*]
\vspace{-2mm}
\item \textbf{Zero-Shot (ZS)}. The pre-trained backbones are directly applied for V $\leftrightarrow$ T retrieval without fine-tuning on downstream datasets. We perform zero-shot retrieval via: $(i)$ \textit{dual encoders,} computing the cosine similarity between video clips and textual narrations, and (ii) \textit{fusion encoder}, incorporating the pre-trained VTM head to compute the video-text matching score. 
\item Task-specific \textbf{Head-tune (HT)}. We extract features using the frozen encoder and train task-specific heads\footnote{VSLNet \cite{zhang2020span} for EgoNLQ, VSGN \cite{zhao2021video} for EgoMQ, single-layer transformer encoder \cite{vaswani2017attention} for summarization, and linear layers for video QA.} using the training split of downstream datasets. 
\item \textbf{Fine-tune (FT)}. We fine-tune the entire pre-trained video-text model end-to-end using the training split of downstream datasets.
\end{itemize}

\subsection{Implementation Details}

\begin{table}[!t]
\centering
\small
\setlength{\tabcolsep}{4pt}
\resizebox{0.9\columnwidth}{!}{\begin{tabular}{@{}l | c c c c c @{}}
\toprule
\multirow{1}{*}{\bf Method} & \multicolumn{1}{c }{\bf Video-$1$} & \multicolumn{1}{c }{\bf Video-$2$} & \multicolumn{1}{c }{\bf Video-$3$} & \multicolumn{1}{c }{\bf Video-$4$} & \multicolumn{1}{c}{\bf Average} \\ 



\midrule
SeqDPP \cite{gong2014diverse} & 36.59 & 43.67 & 25.26 & 18.15 & 30.92 \\
SH-DPP \cite{sharghi2016query} & 35.67 & 42.72 & 36.51 & 18.62 & 33.38 \\
QC-DPP \cite{sharghi2017query} & 48.68 & 41.66 & 36.51 & 29.96 & 44.19 \\
TPAN \cite{zhang2018query} & 48.74 & 45.30 & 56.51 & 33.64 & 46.05 \\
CHAN \cite{xiao2020convolutional} & 49.14 & 46.53 & 58.65 & 33.42 & 46.94 \\
HVN \cite{jiang2019hierarchical} & \underline{51.45} & 47.49 & 61.08 & 35.47 & 48.87 \\
QSAN \cite{xiao2020query} & 48.52 & 46.64 & 56.93 & 34.25 & 46.59 \\
WHM \cite{nalla2020watch} & 50.96 & 48.28 & 58.41 & \textbf{39.18} & 49.20 \\
IntentVizor \cite{wu2022intentvizor} & 51.27 & 53.48 & \underline{61.58} & 37.25 & \underline{50.90} \\
\midrule
EgoVLP$^{\dagger}$ \cite{linegocentric}& 49.64 & \underline{53.60} & 59.87 & 35.76 & 49.72 \\
\rowcolor{Light}
\model\ & \bf 53.30 & \bf 54.13 & \bf 62.64 & \underline{38.25} & \bf 52.08\\

\midrule

\bf \textcolor{blue}{$\Delta_{\text{Ours - EgoVLP}}$} & \textcolor{blue}{3.66} \textcolor{blue}{$\uparrow$} & \textcolor{blue}{0.53} \textcolor{blue}{$\uparrow$} & \textcolor{blue}{2.77} \textcolor{blue}{$\uparrow$} & \textcolor{blue}{2.49} \textcolor{blue}{$\uparrow$} & \textcolor{blue}{2.36} \textcolor{blue}{$\uparrow$}\\

\bottomrule
\end{tabular}}
\captionof{table}{\textbf{Performance on query-focused video summarization (QFVS).} Existing baselines are trained end-to-end, whereas \model\ only learns a tiny head on top of pre-trained encoders. $^{\dagger}$EgoVLP denotes the performance achieved by the officially released checkpoint.  }\label{tab:main_qfvs}
\vspace{-2mm}
\end{table}

We use TimeSformer-B \cite{bertasius2021space, bain2021frozen} and RoBERTa-B \cite{liu2019roberta} as our video and language backbones. The video encoder has 12 layers and 12 heads, and is configured with the patch size of $16 \times 16$ and the hidden dimension of $768$. The spatial attention modules are initialized from a ViT \cite{dosovitskiy2021an}.
We resize videos to $224 \times 224$ and sample $4$ frames per video for pre-training and $16$ for fine-tuning on downstream tasks. We use RoBERTa-B pre-trained on English Wikipedia and Toronto Book Corpus. For our best model,\footnote{An ablation on the number of fusion layers is provided in Section \ref{sec:ablation_study}.} we fuse the top $6$ layers of the two encoders. We pre-train our model for $20$ epochs with a batch size of $256$, using AdamW \cite{loshchilovdecoupled} with a peak learning rate of $3$e-$5$ for the backbones and $12$e-$5$ for the cross-modal parameters. We use linear warmup over the first $2$ epochs and use linear decay. Pre-training takes five days on $32$ A$100$ GPUs. Other necessary pre-training and downstream details are given in the supplementary.



\subsection{Main Results} \label{sec:main_results}

We use \textbf{boldface} and \underline{underline} for the best and second-best performing methods in every table and indicate the performance improvements over the state-of-the-art with \textcolor{blue}{$\Delta$}. 

\vspace{1mm}

\noindent \textbf{Ego4D:} Table \ref{tab:egomcq_nlq} and \ref{tab:egomq} present the performance of \model\ on three different Ego4D benchmarks: EgoMCQ, EgoNLQ and EgoMQ. On EgoMCQ, our model achieves $91.0\%$ inter-video and $60.9\%$ intra-video accuracy, significantly improving over the baselines. Note that \model\ achieves $1\%$ absolute gain on the challenging intra-video MCQ task over \lavila, which is trained using $15\times$ more narrations generated by a pre-trained large language model, GPT-$2$ \cite{radford2019language}. On EgoNLQ, \model\ yields an impressive gain of $2.11\%$ R@$1$ for IoU = $0.3$ over EgoVLP. Moreover, using a smaller task-specific head and fewer epochs of head-tuning, \model\ outperforms existing baselines, which indicates the importance of learning cross-modal information during pre-training.\footnote{Additional details are provided in supplementary.} On the uni-modal grounding task, EgoMQ, our framework also sets a new state-of-the-art, outperforming EgoVLP by $1.54\%$ R@$1$ for IoU = $0.3$, implying the flexibility of \textit{fusion in the backbone} over dual and shared encoder-based pre-training. 

\vspace{1mm}

\begin{table}[!t]
\centering

\small
\setlength{\tabcolsep}{4pt}
\resizebox{0.97\columnwidth}{!}{\begin{tabular}{l c | c c c | c c c }

\toprule

\multirow{2}{*}{\bf Method} & \multirow{2}{*}{\bf \centering Eval.}  & \multicolumn{3}{c |}{\bf Direct} & \multicolumn{3}{c}{\bf Indirect}  \\ 

& & Open & Binary & All &  Open & Binary & All \\

\midrule



VisualBERT \cite{li2019visualbert} & FT & 24.62 & 68.08 & 37.93 & 21.05 & 57.61 & 37.01\\
PSAC \cite{li2019beyond} & FT & 26.97 & 65.95 & 38.90 & 15.31 & 57.75 & 32.72 \\
HME \cite{fan2019heterogeneous} & FT & 27.66 & 68.60 & 40.16 & 18.27 & 52.55 & 33.06 \\
HGA \cite{jiang2020reasoning} & FT & 22.75 & 68.53 & 36.77 & 8.66 & 53.72 & 28.36 \\
HCRN \cite{le2020hierarchical} & FT & 30.23 & 69.42 & 42.40 & \underline{27.82} & \underline{59.29} & \underline{41.56} \\
ClipBERT \cite{lei2021less} & FT & 27.70 & 67.52 & 39.87 & 11.17 & 40.71 & 24.08 \\

\midrule 

EgoVLP$^{\dagger}$ \cite{linegocentric} & FT & \underline{31.69} & \underline{71.26} & \underline{42.51} & 27.04 & 55.28 & 38.69 \\
\rowcolor{Light}
\model\ & FT & \bf 35.56 & \bf 75.60 & \bf 46.26 & \bf 29.14 & \bf 59.68 & \bf 42.28 \\

\midrule 

\bf \textcolor{blue}{$\Delta_{\text{Ours - EgoVLP}}$} & FT & \textcolor{blue}{3.87} \textcolor{blue}{$\uparrow$} & \textcolor{blue}{4.34} \textcolor{blue}{$\uparrow$} & \textcolor{blue}{3.75} \textcolor{blue}{$\uparrow$} & \textcolor{blue}{2.10} \textcolor{blue}{$\uparrow$} & \textcolor{blue}{4.40} \textcolor{blue}{$\uparrow$} & \textcolor{blue}{3.59} \textcolor{blue}{$\uparrow$} \\

\midrule 

EgoVLP$^{\dagger}$ \cite{linegocentric} & HT & \underline{20.52} & \underline{64.63} & \underline{32.76} & \underline{16.87} & \underline{48.40} & \underline{29.19}\\
\rowcolor{Light}
\model\ & HT & \bf 26.59 & \bf 69.10 & \bf 37.87 & \bf 22.11 & \bf 57.19 & \bf 35.20 \\

\midrule

\bf \textcolor{blue}{$\Delta_{\text{Ours - EgoVLP}}$} & HT & \textcolor{blue}{6.07} \textcolor{blue}{$\uparrow$} & \textcolor{blue}{4.47} \textcolor{blue}{$\uparrow$} & \textcolor{blue}{5.11} \textcolor{blue}{$\uparrow$} & \textcolor{blue}{5.24} \textcolor{blue}{$\uparrow$} & \textcolor{blue}{8.79} \textcolor{blue}{$\uparrow$} & \textcolor{blue}{6.01} \textcolor{blue}{$\uparrow$}\\


\bottomrule
\end{tabular}}
\caption{\textbf{Performance on EgoTaskQA \textit{direct} and \textit{indirect} splits.} \model\ outperforms prior work across all
settings, metrics, and data splits. $^{\dagger}$EgoVLP denotes the performance achieved by the officially released checkpoint.}\label{tab:egotaskqa}
\end{table}

\begin{table}[!t]
\centering

\small
\setlength{\tabcolsep}{4pt}
\resizebox{\columnwidth}{!}{\begin{tabular}{l c | c | l c | c c}

\toprule

\multirow{2}{*}{\bf Method} & \multirow{2}{*}{\bf \centering Eval.} & \bf CharadesEgo  & \multirow{2}{*}{\bf Method} & \multirow{2}{*}{\bf \centering Eval.} & \multicolumn{2}{c}{\bf EK-100 MIR}  \\ 

& & mAP & & & mAP & nDCG \\

\midrule 

Actor \cite{sigurdsson2018actor} & FT & 20.0 & S$3$D \cite{xie2018rethinking} & FT & 29.2 & 44.7 \\
SSDA \cite{choi2020unsupervised} & FT & 23.1 & MME \cite{wray2019fine} & FT & 38.5 & 48.5 \\
Ego-Exo \cite{li2021ego} & FT & 30.1 & JPoSE \cite{wray2019fine} & FT & 44.0 & 53.5 \\
EgoVLP \cite{linegocentric} & FT & 32.1 & EgoVLP \cite{linegocentric} & FT & 45.0 & 59.4\\
HierVL-Avg \cite{ashutosh2023hiervl} & FT & 32.6 & HierVL-Avg \cite{ashutosh2023hiervl} & FT & 44.9 & 59.8 \\
HierVL-SA \cite{ashutosh2023hiervl} & FT & \underline{33.8} & HierVL-SA \cite{ashutosh2023hiervl} & FT & \underline{46.7} & \underline{61.1} \\
\rowcolor{Light}
\model\ & FT & \textbf{34.1} & \model\ & FT & \bf 47.3 & \bf 61.9 \\

\midrule 

\bf \textcolor{blue}{$\Delta_{\text{Ours - EgoVLP}}$} & FT & \textcolor{blue}{2.0} \textcolor{blue}{$\uparrow$} & \bf \textcolor{blue}{$\Delta_{\text{Ours - EgoVLP}}$} & FT & \textcolor{blue}{2.3} \textcolor{blue}{$\uparrow$} & \textcolor{blue}{2.5} \textcolor{blue}{$\uparrow$}\\

\bf \textcolor{blue}{$\Delta_{\text{Ours - HierVL-SA}}$} & FT & \textcolor{blue}{0.3} \textcolor{blue}{$\uparrow$} & \bf \textcolor{blue}{$\Delta_{\text{Ours - HierVL-SA}}$} & FT & \textcolor{blue}{0.6} \textcolor{blue}{$\uparrow$} & \textcolor{blue}{0.8} \textcolor{blue}{$\uparrow$}\\

\midrule

EgoVLP \cite{linegocentric} & ZS & 25.0 & EgoVLP \cite{linegocentric} & ZS & 16.6 & 23.1 \\
HierVL-Avg \cite{ashutosh2023hiervl} & ZS & 25.2 & HierVL-Avg \cite{ashutosh2023hiervl} & ZS & 16.7 & 23.5\\
HierVL-SA \cite{ashutosh2023hiervl} & ZS & \underline{26.0} & HierVL-SA \cite{ashutosh2023hiervl} & ZS & \underline{18.9} & \underline{24.7}\\
\rowcolor{Light}
\model\ & ZS & \bf 26.2 & \model\ & ZS & \bf 26.7 & \bf 29.1 \\

\midrule 

\bf \textcolor{blue}{$\Delta_{\text{Ours - EgoVLP}}$} & ZS & \textcolor{blue}{1.2} \textcolor{blue}{$\uparrow$} & \bf \textcolor{blue}{$\Delta_{\text{Ours - EgoVLP}}$} & ZS & \textcolor{blue}{10.1} \textcolor{blue}{$\uparrow$} & \textcolor{blue}{6.0} \textcolor{blue}{$\uparrow$}\\

\bf \textcolor{blue}{$\Delta_{\text{Ours - HierVL-SA}}$} & ZS & \textcolor{blue}{0.2} \textcolor{blue}{$\uparrow$} & \bf \textcolor{blue}{$\Delta_{\text{Ours - HierVL-SA}}$} & ZS & \textcolor{blue}{7.8} \textcolor{blue}{$\uparrow$} & \textcolor{blue}{4.4} \textcolor{blue}{$\uparrow$}\\

\bottomrule
\end{tabular}}
\caption{\textbf{Performance on CharadesEgo and EK-100 MIR.} \model\ achieves significant gains in fine-tuning and zero-shot settings for both tasks. Results are achieved by dual-encoder-based inference.} \label{tab:charades_epic}
\vspace{-2mm}
\end{table}

\noindent \textbf{QFVS:} We evaluate \model\ on query-focused video summarization task. The QFVS dataset contains only $135$ video-query training samples with long ($3$-$5$ hours) videos, and all existing baselines are trained end-to-end. In contrast, we learn a tiny head (single-layer transformer) on top of the pre-trained encoders. As shown in Table \ref{tab:main_qfvs}, our model persistently attains the state-of-the-art F-$1$ score across all four videos in this dataset. The pre-trained video-language representation helps \model\ to achieve strong performance, whereas the baselines struggle to learn good cross-modal features due to the small training set.  

\vspace{1mm}

\noindent \textbf{EgoTaskQA:} Table \ref{tab:egotaskqa} shows the results on the egocentric video question-answering tasks on the EgoTaskQA dataset. Our model achieves significant gains across various baselines in the fine-tuning regime. Notably, \model\ performs consistently well in the challenging \textit{indirect} split, which demonstrates its ability to solve complicated reference tasks. In the head-tuning regime, we only learn a linear layer on top of frozen encoders, where \model\ beats EgoVLP by a strong margin, which proves the efficacy of cross-modal pre-trained representation.  

\vspace{1mm}

\begin{figure*}[!h]
  \centering
\includegraphics[width=0.995\linewidth]{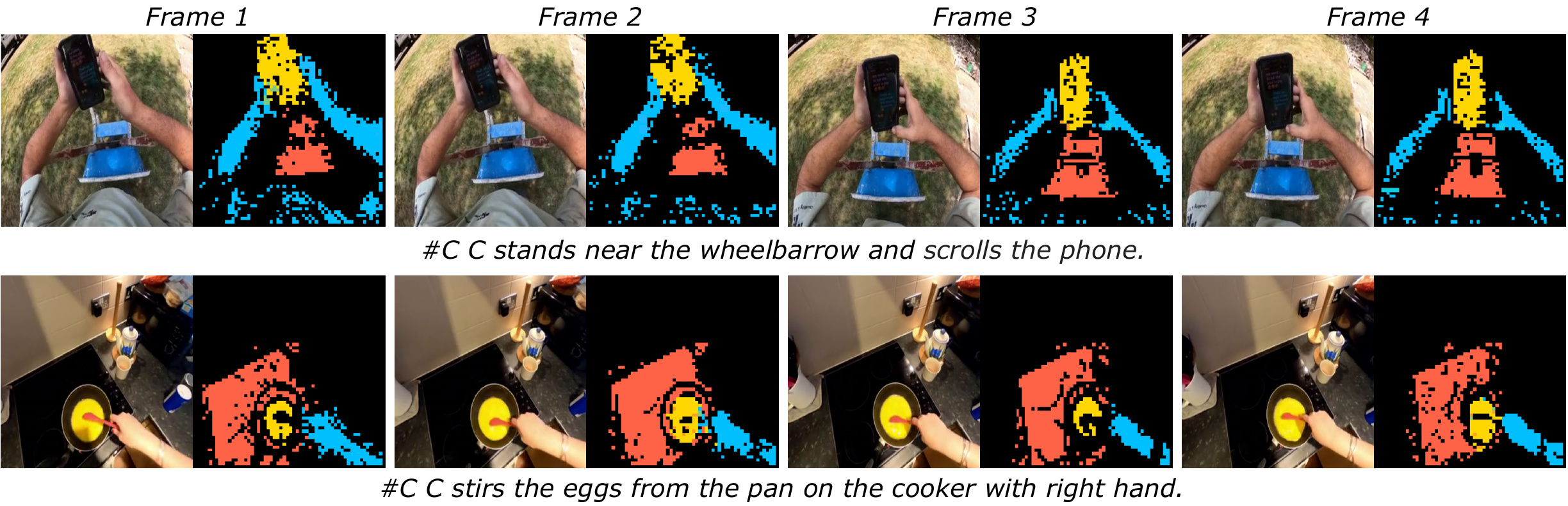}

\caption{\textbf{Text-to-video cross-attention from multiple heads in the last layer of \model\ with $16 \times 16$ patches.} We look at the attention maps of the \texttt{[CLS]} token from the text encoder on input video frames. Different heads, depicted in different colors, focus on different objects or parts. These maps show the strong cross-modal representation learned by \model\ during pre-training, which helps to enhance performance on video-language downstream tasks.
}
\label{fig:visualization_main}
\vspace{-2mm}
\end{figure*}

\begin{table}[!t]
\centering
\small
\setlength{\tabcolsep}{4pt}
\resizebox{0.95\columnwidth}{!}{\begin{tabular}{@{}c c c c | c c @{}}

\toprule

\multirow{2}{1.75cm}{\bf \centering Fusion Strategy} & \multirow{2}{1.5 cm}{\bf \centering \# Fusion Layers} & \multirow{2}{1.5 cm}{\bf \centering \#Trainable Params.} & \multirow{2}{2 cm}{\bf \centering GMACs per instance}  & \multicolumn{2}{c}{\bf EgoMCQ} \\

& & & & Inter & Intra \\

\midrule



\CC{} & \CC{} 3 & \CC{} 374.5M & \CC{} 288.62 & \CC{} 90.5 & \CC{} 60.0\\
\CC{} Fusion in the & \CC{} 6 & \CC{} 381.6M & \CC{} 300.16 & \CC{} \bf 91.0 & \CC{} \bf 60.9\\
\CC{} Backbone & \CC{} 9 & \CC{} 388.7M & \CC{} 311.71 & \CC{} \bf 91.0 & \CC{} \bf 60.9\\
\CC{} & \CC{} 12 & \CC{} 395.8M & \CC{} 323.26 & \CC{} \bf 91.0 & \CC{} \bf 60.9\\

\midrule

\multirow{4}{1.75cm}{\centering Additional Fusion Layers} & 3 & 396.9M & 402.88 & 90.5 & 60.3\\
& 6 & 414.6M & 437.90 & 90.5 & 60.8  \\
& 9 & 432.4M & 472.91 & 90.6 & 60.8\\
& 12 & 450.1M & 507.92 & 90.6 & \bf 60.9\\

\bottomrule

\end{tabular}}
\caption{\textbf{Ablation study on fusion strategies.} Our proposed \textit{fusion in the backbone} strategy performs slightly better than using fusion-specific transformer layers, but with less parameters and less compute .}\label{tab:fusion_ablation}
\vspace{-3mm}
\end{table}

\noindent \textbf{CharadesEgo:} This is a multi-class action recognition task, with class names as short text phrases. We convert this to a video-to-text (V $\to$ T) retrieval problem as in CLIP \cite{radford2021learning}, and perform dual-encoder-based retrieval. As shown in Table \ref{tab:charades_epic}, \model\ obtains a new state-of-the-art in both fine-tuning and zero-shot regimes. Since CharadesEgo videos are significantly different from Ego4D, being captured by crowd-sourced workers using mobile cameras, these results demonstrate the generalizability of \model.

\vspace{1mm}

\noindent \textbf{EK-100:} Table \ref{tab:charades_epic} shows our results on EK-100 MIR. In the fine-tuning regime, \model\ achieves noticeable improvements over the supervised approaches (S$3$D, MME, JPoSE) and VLP methods (EgoVLP, HierVL). In the zero-shot setup, \model\ beats EgoVLP and HierVL by $7.8\%$ mAP and $4.4\%$ nDCG scores. The consistent performance gains again show the quality of pre-trained encoders.

\subsection{Ablation Study} \label{sec:ablation_study}

\noindent \textbf{Fusion in the Backbone:} We compare our fusion module to the conventional practice of using fusion-specific transformer layers, which we implement following ALBEF \cite{li2021align}.\footnote{ \url{https://github.com/salesforce/ALBEF/}} Table \ref{tab:fusion_ablation} shows that the proposed fusion strategy performs slightly better than stacked fusion layers. For both methods, increasing the number of fusion layers to $6$ results in a non-trivial performance gain. However, our proposed architecture is significantly more parameter- and compute-efficient. For instance, with $6$ fusion layers, the proposed architecture contains $33$M fewer parameters and requires $45\%$ lesser computing cost, which shows the efficacy of our method.


\begin{table}[!t]
\centering
\small
\setlength{\tabcolsep}{4pt}
\resizebox{0.995\columnwidth}{!}{\begin{tabular}{@{}c c c c | c c | c c | c c @{}}

\toprule

\multicolumn{4}{c |}{\multirow{2}{*}{\bf Pre-training Objectives}} & \multicolumn{6}{c }{\textbf{EgoMCQ} (\%)}  \\ 

\cmidrule{5-10}

& & & & \multicolumn{2}{c|}{\bf Dual Enc.} & \multicolumn{2}{c|}{\bf Fusion Enc.} & \multicolumn{2}{c}{\bf Ensemble} \\

EgoNCE & MLM & VTM & VTM-Hard & Inter & Intra & Inter & Intra & Inter & Intra \\

\midrule

\ding{51} & $-$ & $-$ & $-$ & 89.5 & 52.6 & $-$ & $-$ & $-$ & $-$  \\
\ding{51} & \ding{51} & $-$ & $-$ & 89.6 & 52.4 & $-$ & $-$ & $-$ & $-$\\
\ding{51} & $-$ & $-$ & \ding{51} & 89.6 & 53.4 & \bf 90.6 & 59.1 & \bf 91.0 & 60.0\\
\ding{51} & \ding{51} & \ding{51} & $-$ & 89.5 & 53.6 & 89.1 & 51.5 & 90.2 & 56.8\\
\rowcolor{Light}
\ding{51} & \ding{51} & $-$ & \ding{51} & \bf 89.8 & \bf 56.7 & \bf 90.6 & \bf 59.6 &  \bf 91.0 & \bf 60.9 \\

\bottomrule
\end{tabular}}
\caption{\textbf{Ablation study on different pre-training objectives of \model.} We evaluate on EgoMCQ using our model either as a dual encoder, as a fusion encoder, or an ensemble of both. Removing any objective leads to a performance drop. The flexibility of the proposed fusion in the backbone module helps us boost retrieval performance using an ensembling strategy.}
\label{tab:loss_ablation_egomcq}
\vspace{-2mm}
\end{table}

\vspace{1mm}
\noindent \textbf{Pre-training Objectives:} We ablate different pre-training objectives and evaluate the pre-trained models on EgoMCQ using \model\ as a \textit{dual} encoder, as a \textit{fusion} encoder, and an ensemble of the two by summing their similarity scores for each video-text pair. As shown in Table \ref{tab:loss_ablation_egomcq}, removing any pre-training objective lead to a performance drop. Specifically, VTM with hard-negative mining is largely beneficial across all three evaluation strategies. Fusion encoder-based evaluation brings significant improvements over dual-encoders; moreover, as EgoMCQ contains only $5$ sentences for every video, both evaluation methods offer similar latency. Ensembling the two yields further $1-2\%$ performance gain for both inter- and intra-video accuracy metrics. 

\subsection{Attention Visualization \& Error Analysis} \label{sec:visualization}

In Figure \ref{fig:visualization_main}, we show that different heads in the cross-modal attention can attend to different semantic regions of the video frames, guided by the narration. We observe that the pre-trained model learns well to recognize a wide variety of objects appearing in egocentric actions, such as indoor furniture, cooking appliances, phones, tablets, car steering, bicycle handles, etc. Such strong cross-modal information learned during pre-training helps \model\ in multi-modal downstream tasks. The visualizations in Figure \ref{fig:visualization_main} are obtained with $960$p video frames, resulting in sequences of $3601$ tokens for $16 \times 16$ patches. However, vastly hindered objects in cluttered environments, especially in low-light conditions, are occasionally not focused. We show such error cases in the supplementary. 
  

\section{Conclusion}

This work introduces \model, the second generation of egocentric video-language pre-training and a significant improvement over the previous generation \cite{linegocentric} by incorporating cross-modal fusion directly into the video and language backbones. Our proposed \textit{fusion in the backbone} strategy is lightweight, compute-efficient, and allows us to unify various VL tasks in a flexible and efficient manner. We conduct extensive experiments to demonstrate the effectiveness of \model\ on a wide range of downstream tasks, consistently achieving state-of-the-art performance. Moreover, we visually demonstrate the effectiveness of the learned cross-attention representation. 

\section*{Acknowledgement}
The codebase for this work is built on the EgoVLP \cite{linegocentric}, \lavila \cite{zhao2022learning}, FIBER \cite{doucoarse}, and VSLNet \cite{zhang2020span} repository. We would like to thank the respective authors for their contribution, and the Meta AI team for discussions and feedback. Shraman Pramanick and Rama Chellappa were partially supported by a MURI program from the Army Research Office under the grant W911NF17-1-0304.

{\small
\bibliographystyle{ieee_fullname}
\bibliography{main}
}

\newpage
\clearpage
\appendix
\counterwithin{figure}{section}
\numberwithin{table}{section}

\section{Radar Chart Figure 1 Details}

Here, we explain the details of the radar chart in Figure \ref{fig:results_summary}, which summarizes the comparative performance of \model\ with EgoVLP \cite{linegocentric}. First, for illustrative purposes, we normalize each axis by the score achieved by \model, which turns the axes in the range $(0, 1]$. Next, we keep the origin of each axis at $0.7$ normalized value, which reasonably separates the inner and outer frames for better readability. Finally, we annotate each vertex with absolute performance metric scores. Notably, in most previous radar chats in the vision-language literature \cite{wang2022image, yu2022coca}, the axes have different scales and shifts,  which may cause misinterpretations and fallacies. However, our illustration is uniform and accurate to scale.      


\section{Algorithm}
The algorithm for pre-training \model\ is given in Algorithm \ref{alg:algo_egovlpv2}. Section \ref{sec:pre_training_objectives} provides details of different pre-training objectives.

\section{Dataset Details}

This section provides additional details of our pre-training and downstream datasets.

\vspace{1mm}

\noindent \textbf{Ego4D \& EgoClip:} Ego4D \cite{grauman2022ego4d} is the first-of-its-kind massive-scale egocentric video-language dataset and benchmark suite. It offers $3670$ hours of daily life activity videos captured by $931$ unique camera wearers from $74$ worldwide locations and $9$ different countries. The videos in Ego4D span hundreds of scenarios (kitchen, laboratory, workshop, porch, shopping, driving, leisure, etc.) with various daytime and weather conditions. A portion of the dataset is accompanied by audio, 3D meshes of the environment, eye gaze, stereo, and synchronized videos from multiple egocentric cameras at the same event. Each narration in Ego4D is a free-form sentence and has a single timestamp. For example, the narration ``\texttt{\#C C walks towards a laundry machine}'' is associated with the video content, which occurs at 28.3\textit{s} of a particular video. However, an activity occurs for a certain duration, and such a single timestamp can not reflect the start and end points where the particular activity takes place. EgoClip \cite{linegocentric} offers a filtered version of Ego4D and designs a contextual variable-length clip pairing strategy to assign every narration with start and end timestamps. Moreover, EgoClip excludes videos that belong to the validation and test sets of the Ego4D benchmark challenges and retains textual annotation from multiple narrators, allowing us to have narration diversity during pre-training. Overall, EgoClip contains $2927$ hours of videos which form $3.8$M clip-text pairs, with an average clip length of $1.0$\textit{s} and a standard deviation of $0.9$\textit{s}. We use this EgoClip version of Ego4D for pre-training. We evaluate \model\ on three different downstream benchmarks of Ego4D: multiple-choice questions (EgoMCQ), natural language query (EgoNLQ), and moment query (EgoMQ).  


\begin{algorithm}[!t]\small
	\caption{Pre-training \model}
	\label{alg:algo_egovlpv2}
	\begin{algorithmic}
		\Require Batch $\mathcal{B}_{N}:\{x_{vid},x_{text}\}$ \\
        Learnable gating parameter: $\alpha$\\
        \model\ Encoder: $\mathcal{F} : \begin{cases} \mathcal{F}_{\mathrm{dual}} \; \text{if} \; \alpha = 0 \\ \mathcal{F}_{\mathrm{fused}} \; \text{if} \; \alpha \neq 0  \end{cases}$
		\For{$(x_{vid}, x_{text})\in\mathcal{B}_{N}$}
		\State $\mathcal{L}_{\mathrm{EgoNCE}} \gets {\mathrm{EgoNCE}}(\mathcal{F}_\mathrm{dual}(x_{vid},x_{text}))$ \Comment{EgoNCE}

        \State $x_{text}^{\text{MLM}} \gets Mask(x_{text})$
		\State $\mathcal{L}_{\mathrm{MLM}} \gets \text{MLM}(\mathcal{F}_{\mathrm{fused}}(x_{vid},x_{text}^{\text{MLM}}))$ \Comment{MLM}

        \State $x_{text}^{\text{VTM}} \gets HardNeg(x_{text})$ 
		\State $\mathcal{L}_{\mathrm{VTM}} \gets \text{VTM}(\mathcal{F}_{\mathrm{fused}}(x_{vid},x_{text}^{\text{VTM}}))$ \Comment{VTM}
        \State $\mathcal{L}_{\mathrm{total}}\gets(1-\gamma-\delta) \mathcal{L}_\mathrm{EgoNCE} + \gamma \mathcal{L}_\mathrm{MLM} + \delta \mathcal{L}_\mathrm{VTM}$
		\EndFor
		\State Back-prop into $\mathcal{F}$ end-to-end with $\mathcal{L}_{\mathrm{total}}$.
	\end{algorithmic}

\end{algorithm}

\vspace{1mm}

\noindent \textbf{QFVS:} The query-focused video summarization (QFVS) \cite{sharghi2017query} dataset builds upon previously existing UT egocentric (UTE) \cite{lee2012discovering} dataset, which contains four $3$-$5$ hours long videos captured in uncontrolled everyday scenarios. QFVS curates $46$ queries for every video, where each query contains two distinct concepts (nouns) \cite{yeung2014videoset, sharghi2016query, borth2013sentibank}. For example, a query can be \{HAT, PHONE\}, or \{FOOD, DRINK\}. These $46$ queries cover four distinct scenarios: $(i)$ both the concepts appear in the same video shot ($15$ such queries),\footnote{QFVS defines every consecutive 5\textit{s} video clip as a shot.} $(ii)$ the concepts appear in the video, but not in a single shot ($15$ such queries), $(iii)$ only one concept appears in the video ($15$ such queries), and $(iv)$ none of the concepts in the query are present in the video ($1$ such query). We use prompt engineering to generate natural language using the concepts in the query and feed the sentence in our model. For instance, a given query \{HAT, PHONE\} is converted as ``\textit{All scenes containing hats and phones}''. We use $10$ different prompts during head-tuning. The QFVS dataset also annotates concepts for every video shot. It proposes a robust evaluation strategy: find the similarity between the concepts in the generated and ground truth summary by maximum weight matching of a bipartite graph, and compute precision, recall, and F1 score from the number of matched concepts. This evaluation strategy helps to capture how well a system summary can retain semantic information instead of visual quantities, as used in previously existing evaluation methods, such as a system-generated summary has to consist of the same key units (frame or shot) as in the user summary \cite{chu2015video, song2015tvsum, xu2015gaze} or comparing pixels and low-level features \cite{gong2014diverse, khosla2013large, kim2014joint, zhang2016summary, zhao2014quasi}. 

\vspace{1mm}

\noindent \textbf{EgoTaskQA:} The EgoTaskQA \cite{jiaegotaskqa} benchmark uses the same egocentric videos as the LEMMA dataset \cite{jia2020lemma}, which contains goal-oriented and multi-tasked human activities with rich human-object interactions and action dependencies in both single-agent and two-agent collaboration scenarios. The videos are segmented into clips with an average duration of $25$\textit{s}. The questions in the EgoTaskQA dataset are machine-generated and aim to evaluate models' capabilities to describe, explain, anticipate, and make counterfactual predictions about goal-oriented events. The answers are of two types - open-answer queries and binary statement verifications. The EgoTaskQA dataset contains $40$K balanced question-answer pairs selected from $368$K programmatically generated questions from $2$K egocentric videos. Moreover, this dataset offers two different benchmark splits $(i)$ \textit{normal} or \textit{direct} split where the train, test, and validation sets are randomly sampled in a $3$:$1$:$1$ ratio and $(ii)$ \textit{indirect} split where the actions and objects are strongly correlated and test the model's task understanding capability with challenging questions. We approach the video QA as a classification task and report accuracy for open queries and binary verification in the direct and indirect splits. 

\vspace{1mm}

\noindent \textbf{CharadesEgo:} The CharadesEgo \cite{sigurdsson2018charades} dataset consists of $68.5$K annotated samples from $7860$ videos from both first and third-person views, covering $157$ classes of daily indoor activities. We only use the first-person subset, which contains $3085$ videos for training and $846$ videos for testing. ChardesEgo is originally a multi-class classification problem, with class labels being short phrases like `\textit{Putting something on the shelf.}' We treat this problem to a video-to-text (V $\to$ T) retrieval task as in CLIP \cite{radford2021learning} by leveraging the text encoder to extract features from class names. We directly evaluate the model on the validation set in the zero-shot setting. In the fine-tuning setting, we leverage the $33.1$K training samples to perform an end-to-end fine-tuning of \model. Following the previous literature \cite{linegocentric, zhao2022learning, ashutosh2023hiervl}, we report video-level mAP as the evaluation metric.

\vspace{1mm}

\noindent \textbf{EK-100:} The Epic-Kitchens-100 \cite{damen2022rescaling} dataset contains $100$ hours of egocentric cooking videos. The training set consists of $67.2$K video samples, whereas the validation and test set has $9.6$K and $13.1$K samples, respectively. Each sample is associated with text narration. We perform multi-instance retrieval (V $\leftrightarrow$ T) on the EK-100 dataset, which is challenging due to the significant semantic overlap between different narrations. The evaluation metrics are mean Average Precision (mAP) and the normalized Discounted Cumulative Gain (nDCG).

\section{Implementation Details}

\subsection{Pre-training on EgoClip}

Table \ref{tab:hyperparams} presents the hyper-parameters used during pre-training. We use TimeSformer-B \cite{bertasius2021space, bain2021frozen} and RoBERTa-B \cite{liu2019roberta} as our video and language backbones. We chose the best learning rate using a grid search. We ablate our other design choices in Section \ref{sec:additional_ablation_pretraining}. We use PyTorch’s native FP$16$ mixed precision training and gradient checkpoint during pre-training. 

After every epoch, we validate the pre-trained checkpoint on EgoMCQ and select the model with the best EgoMCQ intra-video score for other downstream tasks. We extract $4$ frames for every video sample during pre-training and reshape those to $224 \times 224$. We also apply standard \texttt{RandomResizedCrop}, \texttt{RandomHorizontalFlip}, \texttt{ColorJitter} and normalization to every frame. We tokenize the text using RoBERTa tokenizer and pad/truncate every narration to a maximum length of $30$. Pre-training takes five days on $32$ A$100$ GPUs. 

\begin{table}[!t]
\centering
  \resizebox{0.9\columnwidth}{!}{\begin{tabular}{@{} l|c|c @{}}
    \toprule
    \bf Hyper-parameters & \bf Notation & \bf Value\\
    \midrule
    \multicolumn{3}{c}{Model}\\
    \midrule
    Video encoder & $-$ & TimeSFormer-B \cite{bertasius2021space, bain2021frozen} \\
    Text encoder & $-$ & $\texttt{roberta-base}$ \cite{liu2019roberta} \\
    Video \& text embedding & $-$ & $768$ \\
    Video encoder patch size & $-$ & $16 \times 16$ \\
    Video \& text projector & $-$ & $4096$-$4096$-$4096$ \\
    \# Fusion layers & $-$ & $6$\\
    \midrule
    \multicolumn{3}{c}{Pre-training}\\
    \midrule
    Batch size & $-$ & $256$\\
    Epochs & $-$ & $20$ \\
    Number of frames & $-$ & $4$ \\
    Frame resolution &  $-$ & $224 \times 224$ \\
    Vocab size & $-$ & $50265$\\
    MLM prob. & $-$ & $0.15$\\
    Max. length of text & $-$ & $30$ \\
    Temp. in Equation 4 & $\tau$ & $0.05$ \\
    MLM \& VTM loss weights & $\gamma$, $\delta$ & $0.25$, $0.5$ \\
    Optimizer & $-$ & AdamW \cite{loshchilovdecoupled} \\
    Peak LR for backbones & $-$ & $3$e$-5$\\
    Peak LR for cross-att & $-$ & $12$e$-5$\\
    Peak LR for loss heads & $-$ & $12$e$-5$\\
    Warmup & $-$ & Linear (first $2$ epochs) \\
    LR decay & $-$ & Linear \\
    End LR & $-$ & $1$e$-7$ \\
    Betas in AdamW & $(\beta_{1}$, $\beta_{2})$ & $(0.9, 0.98)$ \\
    Eps in AdamW & $-$ & $1$e$-8$\\
    Weight decay & $-$ & $1$e$-2$ \\
    
    \bottomrule
  
  \end{tabular}}
  \caption{\textbf{Pre-training hyper-parameter details of \model.}}
  \label{tab:hyperparams}
  \vspace{-2mm}
\end{table}

\subsection{Downstream Settings}

This section presents our fine-tuning and head-tuning strategy for different downstream tasks. For a fair comparison with the baselines \cite{linegocentric, zhao2022learning, ashutosh2023hiervl}, we follow the same downstream configuration as the baselines when possible. The downstream is performed with $16$ frames per video sample. 

\vspace{1mm}

\begin{figure}[!t]
\centering
\includegraphics[scale=0.8]{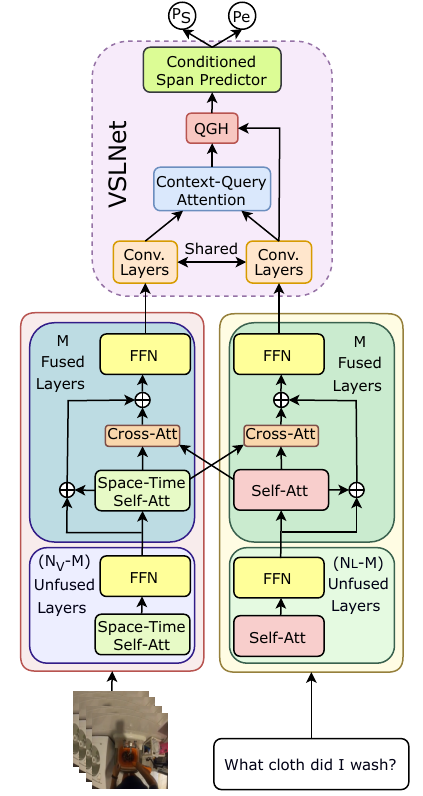}
\caption{\textbf{Entire pipeline for EgoNLQ.} Following EgoVLP \cite{linegocentric} and \lavila\ \cite{zhao2022learning}, we pre-extract video-text features using pre-trained \model, and train VSLNet \cite{zhang2020span} on top of frozen encoders. }
\label{fig:nlq_pipeline}
\vspace{-2mm}
\end{figure}

\noindent \textbf{EgoNLQ:} This task is a video-text localization problem, with each video clip longing up to $1200$\textit{s}. Hence, performing end-to-end fine-tuning can be hard on EgoNLQ. Following \cite{linegocentric, zhao2022learning}, we pre-extract features from the video-text samples using our pre-trained model and train VSLNet \cite{zhang2020span} for $100$ epochs, with a learning rate of $1$e$-3$ and batch size of $32$. We keep all other configurations the same as \cite{linegocentric}.\footnote{\url{https://github.com/showlab/EgoVLP}} However, we observe that we can beat the baselines using even a smaller task head and fewer epochs of tuning, which we describe in Section \ref{sec:ablation_downstreams}. We show the complete EgoNLQ pipeline in Figure \ref{fig:nlq_pipeline}.  

\vspace{1mm}

\noindent \textbf{EgoMQ:} This is a video-only localization problem, and similar to EgoNLQ, the input videos are very long. Hence, end-to-end fine-tuning is also hard to perform on EgoMQ. Following EgoVLP \cite{linegocentric}, we pre-extract video features using pre-trained \model\ and train VSGN \cite{zhao2021video} for $100$ epochs, with a learning rate of $1$e$-4$ and batch size of $32$. We keep all other configurations the same as \cite{linegocentric}. We perform a grid search for other hyper-parameters of VSGN. 

\vspace{1mm}

\noindent \textbf{QFVS:} Query-focused video summarization aims to generate an abridged version of input video guided by a natural language query. To the best of our knowledge, we are the first to unify QFVS as a downstream of a VLP framework. The input videos for this task are very long ($3$-$5$ hours). We first use the unfused $N-M$ layers\footnote{For simplicity, we keep the number of unfused and fused layers the same in the video and text encoder.} of our video and text encoders to extract uni-modal features from every 5-second clip and the text query. Next, we apply the KTS shot boundary detector \cite{potapov2014category} to segment the long video.\footnote{Segmentation helps in two ways: $(i)$ TimeSformer can not process the whole $3$-$5$ hours long video (containing tens of thousands of frames) at once. $(ii)$ Segmentation is also used to convert frame-level prediction scores into key shots. For details, please refer to \cite{ sharghi2017query, fajtl2019summarizing, zhang2016video}.} After this, the query and segment-wise clip features are fed into the top $M$ fused layers of \model\ to compute the multi-modal representation. Finally, we learn an additional single-layer transformer to design the interrelation across all $5$ second long clips in every segment. We train the single-layer transformer for $20$ epochs, with a batch size of $20$, a peak learning rate of $1$e$-5$ using AdamW \cite{loshchilovdecoupled} optimizer, cosine scheduler, and a linear warmup for the first $2$ epochs. We also perform an ablation on the single-layer transformer in Section \ref{sec:ablation_downstreams}.

\vspace{1mm}

\noindent \textbf{EgoTaskQA:} We treat the video QA as a classification problem, where we train linear layers on top of the fused feature representation generated by the pre-trained \model. In the fine-tuning setting, we fine-tune the pre-trained model for $36$ epochs with a batch size of $64$, using the AdamW \cite{loshchilovdecoupled} optimizer. We use cosine annealing with $10$\% linear warmup steps, with the peak learning rate of $2$e$-4$ for the direct split and $1$e$-4$ for the indirect split. In the head-tuning setup, we only train the classifier head on top of frozen backbones with the same configuration. 

\vspace{1mm}

\noindent \textbf{CharadesEgo:} Following \cite{linegocentric, zhao2022learning, ashutosh2023hiervl}, we convert CharadesEgo as a retrieval problem. In the zero-shot setup, we perform dual-encoder-based inference. In the fine-tuning setup, we use EgoNCE as our objective. We fine-tune the model for $10$ epochs with a batch size of $128$ using AdamW \cite{loshchilovdecoupled} optimizer with $(\beta_1, \beta_2) = (0.9, 0.98)$, and weight decay of $0.01$. We use cosine annealing with warmup, with $10$\% linear warmup steps, peak learning rate of $1.5$e$-4$ and end learning rate of $1$e$-7$. Since this is a multi-class dataset, where each video can include multiple actions, we report mAP as the evaluation metric. For input, we sample $16$ frames from each video clip, and reshape the frames into $224 \times 224$.

\vspace{1mm}

\noindent \textbf{EK-100 MIR:} Since a narration can jointly be associated with multiple videos for EK-$100$ multi-instance retrieval task, we use the adaptive multi-instance max-margin loss \cite{wray2019fine} for this task with a margin value of $0.2$. We keep the zero-shot configuration the same as CharadesEgo. We fine-tune the model for $100$ epochs with a batch size of $128$ using AdamW \cite{loshchilovdecoupled} optimizer with $(\beta_1, \beta_2) = (0.9, 0.98)$, and weight decay of $0.01$. We use cosine annealing with warmup, with $10$\% linear warmup steps, peak learning rate of $2$e$-4$ and end learning rate of $1$e$-7$.

\section{Additional Ablations on Pre-training} \label{sec:additional_ablation_pretraining}

\begin{figure}[!t]
  \centering
  \includegraphics[scale=0.55]{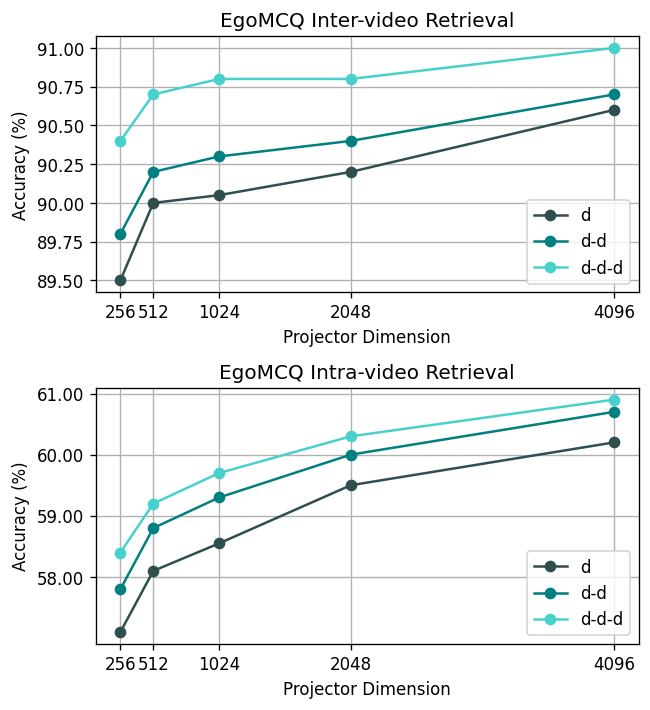}
  \caption{\textbf{Ablation on the projector dimension used in the EgoNCE head.} A $3$-layer projector works better than a single-layer projector. Moreover, an increase in the width of the projector also helps in performance.}
  \label{fig:ablation_supp_projector}
  \vspace{-2mm}
\end{figure}

We conduct additional ablation experiments in this section to validate our design choices. Reported results on EgoMCQ in Table \ref{tab:albation_supp_egonce}, \ref{tab:albation_supp_alpha}, \ref{tab:ablation_supp_batch_frame} and Figure \ref{fig:ablation_supp_projector} are achieved by directly ensembling dual- and fusion-encoder-based inference.

\begin{table}[!t]
\centering
\small
\setlength{\tabcolsep}{4pt}
\resizebox{0.9\columnwidth}{!}{\begin{tabular}{@{}l | c c | c c @{}}

\toprule

\multirow{2}{*}{\bf Pre-training Objectives} & \multicolumn{2}{c |}{\bf EgoNCE Sampling} & \multicolumn{2}{c}{\textbf{EgoMCQ} (\%)}  \\ 

& \quad Pos. & Neg. & Inter & Intra \\

\midrule

InfoNCE + MLM + VTM & \quad $-$ & $-$ & 90.0 & 55.2 \\
EgoNCE + MLM + VTM & \quad \ding{51} & \ding{55} & 90.4 & 58.8\\
EgoNCE + MLM + VTM & \quad \ding{55} & \ding{51} & 90.5 & 59.1\\
\rowcolor{Light}
EgoNCE + MLM + VTM & \quad \ding{51} & \ding{51} & \bf 91.0 & \bf 60.9 \\

\bottomrule
\end{tabular}}
\caption{\textbf{Ablation on EgoNCE sampling strategy.} EgoNCE \cite{linegocentric} helps in improving the performance significantly compared to InfoNCE \cite{oord2018representation}. We also observe that both the positive and negative sampling of EgoNCE is important, and removing any of those leads to a performance drop.}
\label{tab:albation_supp_egonce}
\vspace{-3mm}
\end{table}

\begin{table}[!t]
\centering
\small
\setlength{\tabcolsep}{4pt}
\resizebox{0.4\columnwidth}{!}{\begin{tabular}{@{} l | c c @{}}

\toprule

\multirow{2}{*}{\bf Cross-Att} & \multicolumn{2}{c}{\textbf{EgoMCQ} (\%)}  \\ 

& Inter & Intra \\

\midrule

$\alpha$ = 0.1 & 90.1 & 59.8 \\
$\alpha$ = 0.25 & 90.4 & 59.9\\
$\alpha$ = 0.5 & 90.1 & 58.0 \\
$\alpha$ = 1 & 89.4 & 56.9 \\

\midrule
\rowcolor{Light}
Learnable $\alpha$ & \bf 91.0 & \bf 60.9 \\

\bottomrule
\end{tabular}}
\caption{\textbf{Ablation on the gated cross-attention.} Learnable gating scaler $\alpha$ performs better than a fixed value.}
\label{tab:albation_supp_alpha}
\vspace{-2mm}
\end{table}

\vspace{1mm}

\noindent \textbf{Effect of EgoNCE:} We study the effect of the EgoNCE loss \cite{linegocentric} compared to the more popular InfoNCE objective \cite{oord2018representation}. Given a batch of $N$ video-text pairs, InfoNCE treats the matched $N$ pairs as positives and every other pair as negatives. However, egocentric videos pose two unique challenges: $(i)$ \textcolor{brown}{Same actions} in \textcolor{blue}{different scenarios} appear to be visually different (\textcolor{brown}{\textit{talking on the phone}} \textcolor{blue}{\textit{indoors}} and \textcolor{blue}{\textit{outdoors}}). $(ii)$ \textcolor{brown}{Different actions} in \textcolor{blue}{same scenarios} appear to be similar (\textcolor{brown}{\textit{writing on a tablet}} and \textcolor{brown}{\textit{watching a movie on a tablet}} are visually \textcolor{black}{indistinguishable}). To overcome these challenges, EgoNCE is built upon InfoNCE with two modifications: $(i)$ Besides the matched video-text samples in every batch, all narration pairs which share at least one noun and one verb are treated as positives. $(ii)$ Every batch of $N$ video-text pairs is augmented with another $N$ visually similar videos, often containing different actions in the same scenarios. These added videos with the same texts as in the original batch are treated as additional negatives.

Table \ref{tab:albation_supp_egonce} shows the effect of the modified positive and negative sampling of EgoNCE on \model. First, we observe that replacing EgoNCE with InfoNCE leads to a performance drop of $5.7\%$ accuracy on the challenging intra-video metric of EgoMCQ. Further, discarding either positive or negative sampling also drops the results by $2.1$-$1.8$\% intra-video accuracy. These results align with the findings in \cite{linegocentric} and indicate the efficacy of the EgoNCE objective for egocentric video-language pre-training.  

\vspace{1mm}

\noindent \textbf{Effect of Gated Cross-attention:} Next, we study the importance of gated cross-attention modules with learnable gating scalar, $\alpha$. Table \ref{tab:albation_supp_alpha} shows that a fixed value of $\alpha$ leads to a significant performance drop. In our best pre-trained model, we also find that the learned value of $\alpha$ varies in different layers, ranging from $0.05$ to $0.4$.

\vspace{1mm}
\noindent \textbf{Effect of Projector:} We compare different choices of projector dimensions used in the EgoNCE head in Figure \ref{fig:ablation_supp_projector}. We observe that a three-layer projector works better than single and two-layer projectors. For instance, a $4096$-$4096$-$4096$ dimensional projector improves the EgoMCQ intra-video retrieval performance by $0.85$\% over a single $4096$ dimensional projector. Moreover, an increase in the width of the projector also helps in performance. Hence, we use $4096$-$4096$-$4096$ as our default projector. Notably, these results oppose the findings in Zhao et al. \cite{zhao2022learning}, where the authors observe that using $256$-dimension achieves better performance than a $512$ dimensional projector. The reason behind such results is, in contrast to Zhao et al., \cite{zhao2022learning}, who only use InfoNCE, a larger projector helps us both in EgoNCE and VTM objectives by offering a stronger hard-negative sampling.

\begin{table}
\begin{subtable}[c]{0.21\textwidth}
\centering
\small
\setlength{\tabcolsep}{4pt}
\begin{tabular}{@{} c | c c @{}}

\toprule

\multirow{2}{*}{\bf Batch Size} & \multicolumn{2}{c}{\textbf{EgoMCQ} (\%)}  \\ 

& Inter & Intra \\

\midrule
128 & 90.6 & 59.8\\
\rowcolor{Light}
256 & \bf 91.0 & \bf 60.9 \\
512 & \bf91.0 & 60.6\\
1024 & 90.8 & 60.5\\

\bottomrule
\end{tabular}
\subcaption{\textbf{Ablation on batch size.} EgoMCQ performance is best with a batch size of $256$.}
\label{tab:ablation_supp_batchsize}
\end{subtable}
\hspace{0.75em}
\begin{subtable}[c]{0.21\textwidth}
\vspace{3.75mm}
\centering
\small
\setlength{\tabcolsep}{4pt}
\begin{tabular}{@{} c | c c @{}}

\toprule

\multirow{2}{1.9cm}{\bf \centering \# Frames (Pre-training)} & \multicolumn{2}{c}{\textbf{EgoMCQ} (\%)}  \\ 

& Inter & Intra \\

\midrule
2 & 90.1 & 56.7\\
\rowcolor{Light}
4 & 91.0 & 60.9 \\
5 & 91.2 & 61.2\\
6 & 91.4 & 61.5\\

\bottomrule
\end{tabular}
\subcaption{\textbf{Ablation on number of frames.} Increasing frames improve EgoMCQ performance.}
\label{tab:ablation_supp_frames}
\end{subtable}
\caption{\textbf{Ablation on pre-training batch size (a) and the number of frames (b).} A batch size of $256$ produces the best results. Increasing the number of frames helps in a performance gain. For a fair comparison with the baselines \cite{linegocentric, zhao2022learning, ashutosh2023hiervl}, we keep $4$ as our default frame number.}
\label{tab:ablation_supp_batch_frame}
\vspace{-3mm}
\end{table}

\vspace{1mm}
\noindent \textbf{Effect of Batch Size:} Next, we study the effect of pre-training batch size in Table \ref{tab:ablation_supp_batchsize}. The performance improves using a batch size of $256$ over $128$. However, the performance drops if we further increase the batch size to $512$ or $1024$. Therefore, we use $256$ as our default batch size in all other experiments.

\vspace{1mm}
\noindent \textbf{Effect of Number of Frames:} Lastly, we ablate the number of frames per sample during pre-training in Table \ref{tab:ablation_supp_frames}. We see a good improvement in the EgoMCQ performance when the number of frames is increased to $4$. However, after $4$, the performance improvement diminishes. We keep $4$ as our default frame number for a fair comparison with the baselines \cite{linegocentric, zhao2022learning, ashutosh2023hiervl}, who also use $4$ frames per sample during pre-training. 

\section{Ablations on Downstream} \label{sec:ablation_downstreams}

\begin{table}[!t]
\centering
\small
\setlength{\tabcolsep}{4pt}
\resizebox{0.8\columnwidth}{!}{\begin{tabular}{@{}l | c c c c @{}}

\toprule

\multirow{3}{*}{\bf Model + Task head} & \multicolumn{4}{c}{\bf EgoNLQ validation set}\\ 

&  \multicolumn{2}{c}{mIOU@0.3} &  \multicolumn{2}{c}{mIOU@0.5} \\
& R@1 & R@5 & R@1 & R@5 \\ 

\midrule

SlowFast \cite{feichtenhofer2019slowfast} + VSLNet \cite{zhang2020span} & 5.45 & 10.74 & 3.12 & 6.63 \\
EgoVLP \cite{linegocentric} + VSLNet \cite{zhang2020span} & 10.84 & 18.84 & 6.81 & 13.45 \\
\lavila \cite{zhao2022learning} + VSLNet \cite{zhang2020span} & 10.53 & 19.13 & 6.69 & 13.68 \\ 

\midrule

\model\ + Span & 11.08 & 21.27 & 7.05 & 14.29 \\
\model\ + QGH + Span & 11.95 & 22.86 & 7.64 & 15.80 \\
\rowcolor{Light}
\model\ + VSLNet \cite{zhang2020span} & \bf 12.95 & \bf 23.80 & \bf 7.91 & \bf 16.11 \\

\bottomrule
\end{tabular}}
\caption{\textbf{Ablation on task-head for EgoNLQ.} \model\ beats existing models even using a smaller task-head.}
\label{tab:ablation_nlq}
\vspace{-2mm}
\end{table}

\begin{table}[!t]
\centering
\small
\setlength{\tabcolsep}{4pt}
\resizebox{0.98\columnwidth}{!}{\begin{tabular}{@{}l | c c c c c @{}}

\toprule

\multirow{1}{*}{\bf Model + Task head} & \bf Video-$1$ & \bf Video-$2$ & \bf Video-$3$ & \bf Video-$4$ & \bf Average \\

\midrule

\model\ + Linear layers & 50.17 & 50.95 & 59.38 & 34.58 & 48.77\\
\rowcolor{Light}
\model\ + $1$-layer transformer & \bf 54.97 & \bf 55.74 & \bf 64.10 & \bf 40.83 & \bf 53.91\\
\model\ + $2$-layer transformer & 52.78 & 51.98 & 66.80 & 34.10 & 51.42 \\
\model\ + $3$-layer transformer & 51.87 & 52.45 & 63.75 & 35.55 & 50.91 \\

\bottomrule
\end{tabular}}
\caption{\textbf{Ablation on task-head for QFVS.} A single-layer transformer produces better performance than linear layers and multi-layer transformers.}
\label{tab:ablation_qfvs}
\vspace{-3mm}
\end{table}

This section presents an ablation on downstream task-specific heads for EgoNLQ and QFVS.

\vspace{1mm}
\noindent \textbf{EgoNLQ:} Following EgoVLP \cite{linegocentric} and \lavila\ \cite{zhao2022learning}, we use VSLNet \cite{zhang2020span} as the task-head for EgoNLQ. However, since our model learns cross-modal features during pre-training, we observe that we can beat the previous methods by a significant margin even using smaller task heads. As shown in Table \ref{tab:ablation_nlq}, when we only use the conditional span predictor module, which is just a linear layer, we can beat EgoVLP by $2.43$\% R@5 for IoU=$0.3$. Adding the QGH module further helps in improving the performance. Using the whole VSLNet can significantly beat EgoVLP and \lavila\ across all metrics. Moreover, the previous methods train VSLNet for $200$ epochs, whereas we achieve the best performance within $100$ epochs. These results prove the efficacy of the cross-modal pre-trained representation of \model.

\vspace{1mm}
\noindent \textbf{QFVS:} Next, we compare different heads for QFVS in Table \ref{tab:ablation_qfvs}. Notably, this dataset is very small, with only $135$ training samples. We observe that a single-layer transformer head performs better than linear layers and multi-layer transformers. Linear layers can not model temporal relations across different video shots, which a transformer can efficiently do. However, multi-layer transformers overfit this dataset due to the small training set. Hence, we use a single-layer transformer for QFVS.

\section{Error Analysis}

\begin{figure*}[!t]
  \centering
  \includegraphics[width=0.975\linewidth]{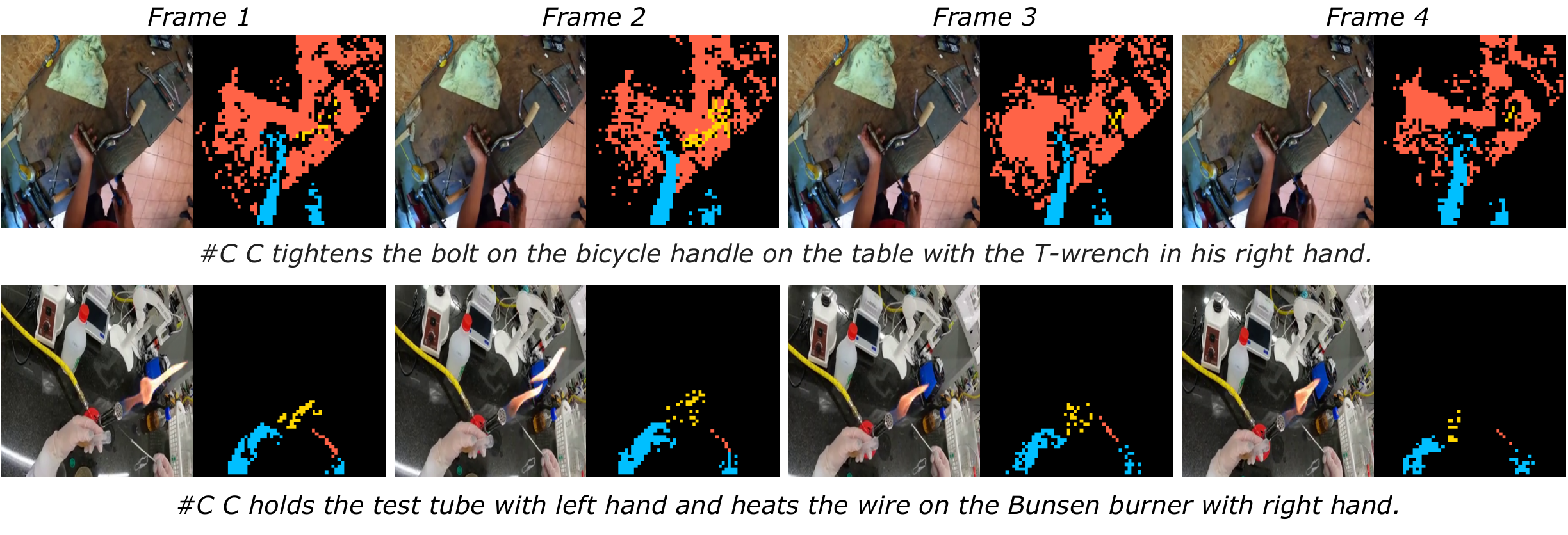}
  
  \caption{\textbf{Limitations of our method:} tiny and hindered objects in cluttered environments are not distinctly attended by the pre-trained \model. We show the attention maps of the \texttt{[CLS]} token from the text encoder on input video frames in the text-to-video cross-attention module of the last layer of \model. Different heads, shown in different colors, focus on various semantic regions of the video frames. The visualizations are obtained with $960$p video frames, resulting in sequences of
  $3601$ tokens for $16 \times 16$ patches.}
  \label{fig:error_analysis}
\end{figure*}

Although \model\ learns impressive cross-modal representation during pre-training, there are still some cases where the model fails to identify tiny and hindered objects, especially in cluttered environments. We show two such examples in Figure \ref{fig:error_analysis}. In the first video, the objects `bicycle handle' and `T-wrench' are barely visible even in human eyes, and thus, \model\ can not consistently attend to these objects in all frames. However, it can recognize larger, more familiar things like tables and human hands. In the second video, we show an egocentric activity in a wet lab, where the camera wearer is wearing gloves, holding a test tube, and heating a wire using a bunsen burner. This is a complex scenario with multi-agent collaborative activities and fine-grained actions. Interestingly, \model\ can correctly identify the human hands and track the motion of the thumb in different frames, even when wearing gloves. However, the test tube and the wire are hindered and are partially attended by the model. Since we pre-train \model\ with $224 \times 224$ video frames, such tiny objects are often hard to be distinguished. However, higher-resolution frames will be more helpful in addressing such intricate scenarios, which we plan to explore in future works. 

\section{Qualitative Downstream Performance}

\begin{figure*}
     \centering
     \begin{subfigure}[b]{0.49\textwidth}
         \centering
         \includegraphics[width=\textwidth]{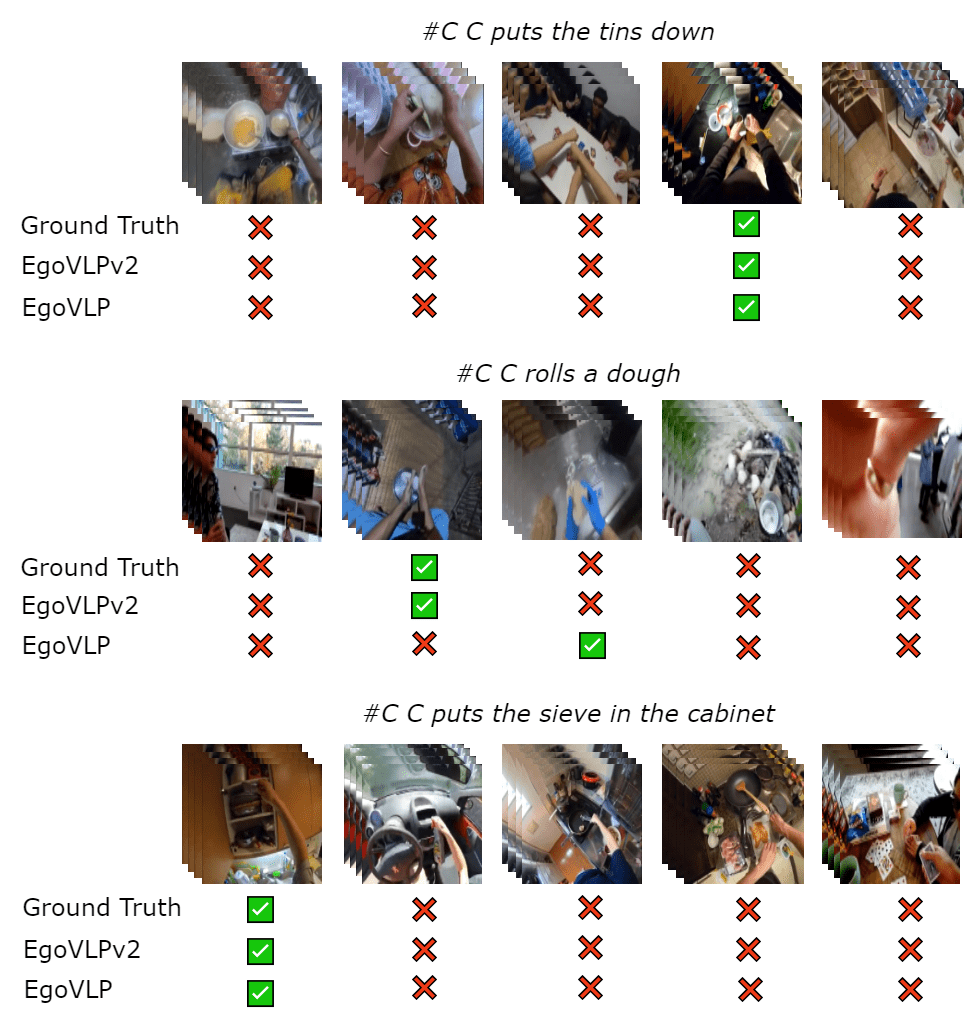}
         \caption{Inter-video MCQ.}
         \label{fig:}
     \end{subfigure}
     \hfill
     \begin{subfigure}[b]{0.49\textwidth}
         \centering
         \includegraphics[width=\textwidth]{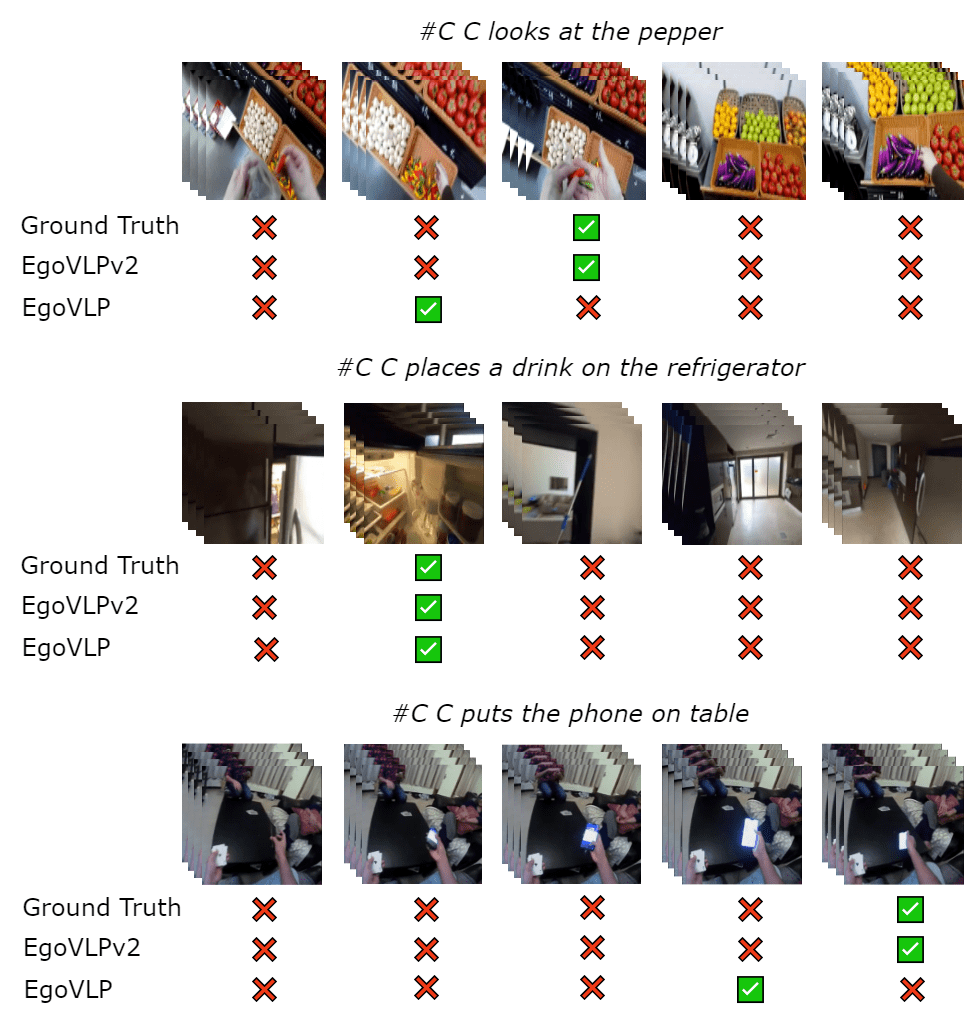}
         \caption{Intra-video MCQ.}
     \end{subfigure}
         \caption{\textbf{Examples of predictions made by EgoVLP \cite{linegocentric} and \model\ on multiple choice questions from EgoMCQ validation set.} \textit{Left:} The ``inter-video'' setting, each question contains $5$ clips from different videos. \textit{Right:} The ``intra-video'' setting, each question contains $5$ contiguous clips from the same video, making it more challenging.}
    \label{fig:egomcq}
\end{figure*}

\vspace{1mm}
\noindent \textbf{EgoMCQ:} In Figure \ref{fig:egomcq}, we show example predictions made by EgoVLP \cite{linegocentric} and \model\ on multiple choice questions from EgoMCQ validation set. \model\ beats EgoVLP substantially on the challenging intra-video setting, where all $5$ choices are visually similar. The VTM head pre-trained with hard-negative sampling helps \model\ to distinguish between similar videos and boosts the performance over EgoVLP.  

\vspace{1mm}

\begin{figure*}[!t]
    \centering
    \includegraphics[width=0.95\textwidth]{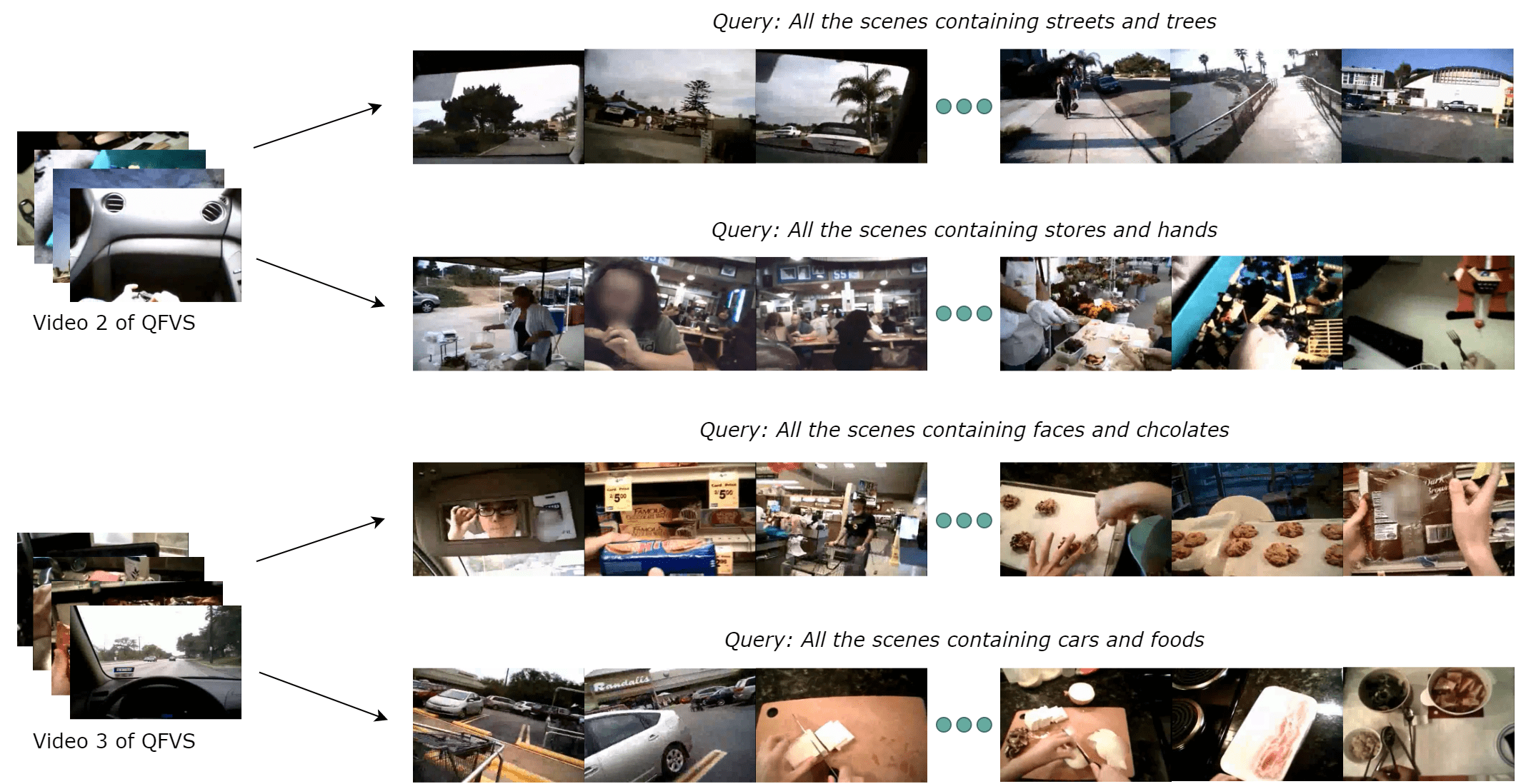}
    \caption{\textbf{Examples of query-focused video summary generated by \model\ on the QFVS daatset.} Given a long egocentric video and a natural language query, the generated summary includes all relevant scenes. For example, the query ``All the scenes containing streets and trees'' summarizes the scenes containing streets and trees in the long input video.}
    \label{fig:qfvs}
\end{figure*}

\noindent \textbf{QFVS:} Figure \ref{fig:qfvs} shows some examples of query-focused summaries generated by \model\ on the QFVS dataset. Given a long egocentric video and a natural language query, our model can summarize all relevant scenes successfully. Notably, the input videos on this dataset are very long ($3$-$5$ hours), and the length of the generated summary is $2$\% input video, which makes this task challenging.

\vspace{1mm}

\noindent \textbf{EgoNLQ:} Figure \ref{fig:nlq} shows examples of predictions made by EgoVLP \cite{linegocentric} and \model\ on text-guided video localization from the EgoNLQ dataset. Given an untrimmed video and a natural language query, this task aims to predict a single temporal window to answer the query. The predictions of \model\ are significantly more aligned with the ground truth than EgoVLP, which supports the impressive quantitative performance gain by \model\ over EgoVLP across all metrics. 

\begin{figure*}[!t]
    \centering
    \includegraphics[scale=0.26]{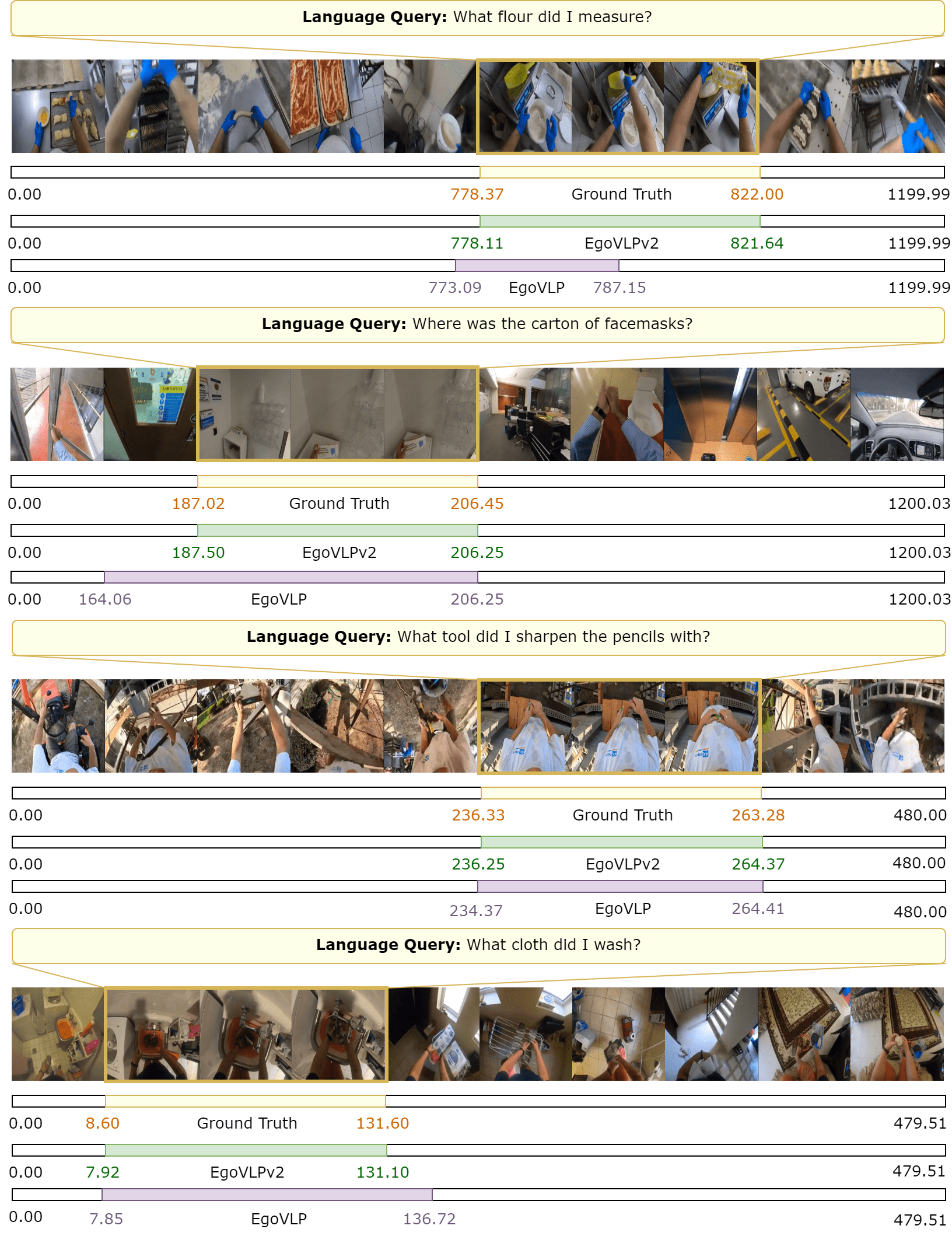}
    \caption{\textbf{Examples of predictions made by EgoVLP \cite{linegocentric} and \model\ on text-guided video localization from the EgoNLQ dataset.} Given an untrimmed video and a language query, the prediction is a single temporal window containing the answer to the query. The predictions of \model\ are significantly more aligned with the ground truth than EgoVLP.}
    \label{fig:nlq}
\end{figure*}

\end{document}